\documentclass{article}
\usepackage{mathrsfs}
\usepackage{amssymb}
\usepackage{amsthm}
\usepackage{amsmath}
\usepackage{enumitem}

\usepackage[a4paper]{geometry}
\usepackage{verbatim}
\usepackage{pdfpages}
\usepackage{graphicx}
\usepackage{setspace}
\usepackage{soul}
\usepackage[numbers]{natbib}
\usepackage{nomencl}

\newtheorem{claim}{Claim}[section]

\providecommand{\makenomenclature}{\makeglossary}
\makenomenclature

\makeatletter
\providecommand{\LyX}{L\kern-.1667em\lower.25em\hbox{Y}\kern-.125emX\@}
\usepackage{hyperref}  
\hypersetup{
     colorlinks=true,
     linkcolor=blue,
     citecolor=blue,
     filecolor=blue,
     urlcolor=blue,
 } 
\usepackage[font={small,bf}, labelfont={small,bf}, margin=1cm]{caption}
\usepackage{titlesec}
\titleformat{\section}{\bfseries\large}{\thesection}{1em}{}
\titleformat{\subsection}{\bfseries\large}{\thesubsection}{1em}{}
\titleformat{\subsubsection}{\bfseries\large}{\thesubsubsection}{1em}{}
\newcommand{\hsp}{\hspace{20pt}}
\usepackage{amsmath}
\usepackage{amssymb}
\usepackage{xcolor}
\titleformat{\chapter}[hang]{\Huge\bfseries}{\thechapter\hsp}{0pt}{\Huge\bfseries}

\title{Feedback-Based Mobile Robot Navigation in 3-D Environments Using Artificial Potential Functions\\
{\large Technical Report}}
\date{}  % empty date
\author{
Ro'i Lang$^{1}$ \and
Elon Rimon$^{1}$
}

\begin{document}
\maketitle
\vspace{-3em}
\begin{center}
$^{1}$Technion -- Israel Institute of Technology, Haifa 3200003, Israel\\
\texttt{roi.lang@campus.technion.ac.il}, \texttt{rimon@technion.ac.il}\\[0.5em]

\end{center}

% ================== Abstract ==================
\vspace{3em}
\begin{abstract}
This technical report presents the construction and analysis of polynomial navigation functions for motion planning in 3-D workspaces populated by spherical and cylindrical obstacles. The workspace is modeled as a bounded spherical region, and obstacles are encoded using smooth polynomial implicit functions. We establish conditions under which the proposed navigation functions admit a unique non-degenerate minimum at the target while avoiding local minima, including in the presence of pairwise intersecting obstacles. Gradient and Hessian analyses are provided, and the theoretical results are validated through numerical simulations in obstacle rich 3-D environments.
\end{abstract}

\newpage
\tableofcontents
\newpage

%%%%%%%%%%%%%%%%%%%%%%%%%%%%%%%%%%%%
\section{Introduction}
\label{sec:intro}
%%%%%%%%%%%%%%%%%%%%%%%%%%%%%%%%%%%%
Robot navigation using artificial potential functions offers a systematic way to integrate global motion planning with real-time feedback control. The central idea is to encode the geometry of the robot \emph{free space}, together with the target location, into a scalar potential function known as a \emph{navigation function}. When used as a feedback control law, the negative gradient of a navigation function guides the robot safely to the target from almost any initial configuration while guaranteeing collision avoidance. As originally introduced in~\cite{koditschek1990robot}, navigation functions are characterized by two fundamental properties:
\begin{enumerate}[topsep=0pt, itemsep=2pt]
\item \textbf{Polarity}: the function admits a unique global minimum at the target configuration, while all other critical points are nondegenerate saddle points.
\item \textbf{Admissibility}: the function attains a uniform maximal value on the boundary of the free space, formed by workspace boundaries and obstacles.
\end{enumerate}

Over the past three decades, navigation functions have been successfully applied to a wide range of robotic systems and environments, including multi-agent coordination~\cite{dimarogonas2012sufficient,hacohen2024navigation}, navigation in dynamic environments~\cite{chen2020navigation,chiang2015path}, mobile manipulation~\cite{tanner2003nonholonomic}, and, more recently, learning-based navigation~\cite{rousseas2021harmonic,rousseas2024reactive}. These developments highlight the value of navigation functions as a unifying framework connecting motion planning, control, and formal guarantees.

Most classical constructions of navigation functions assume that the workspace outer boundary and internal obstacles are convex, and that the free space can be modeled as a so-called \emph{sphere world}, where obstacles are represented by $n$-dimensional spheres. Within this setting, smooth navigation functions with provable properties can be constructed using analytic expressions~\cite{koditschek1990robot,Rimon1988}. When obstacles deviate from spherical geometry, diffeomorphic coordinate transformations are commonly employed to map the free space into an equivalent spherical domain~\cite{rimon1991construction,loizou2012navigation}. While theoretically possible, such transformations are highly environment dependent, computationally expensive, and difficult to extend to general 3-D settings.

\begin{figure}[t!]
    \centering
    \includegraphics[width=0.9\linewidth]{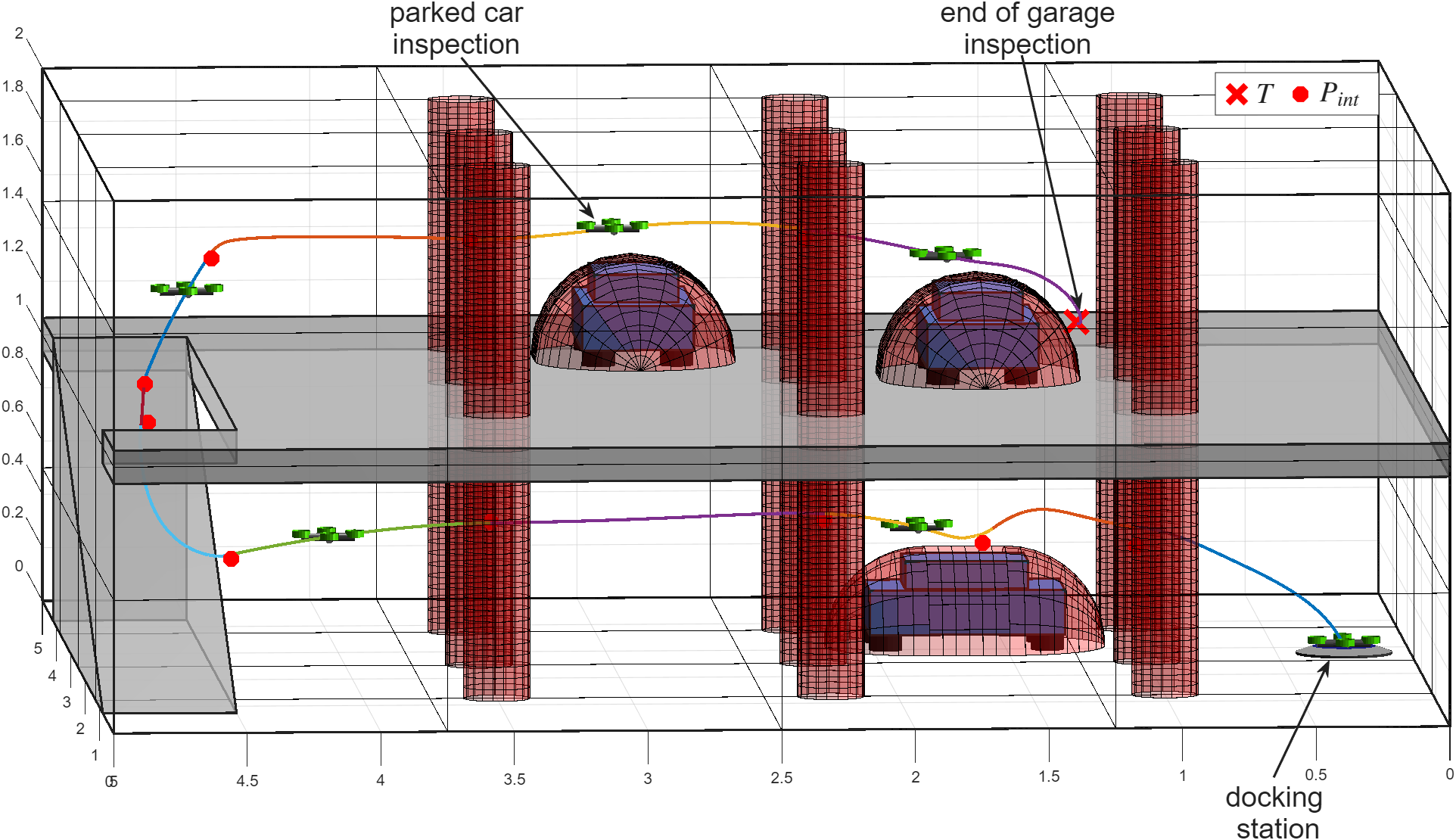}
    \caption{Simulation of a quadrotor navigating in a two-story parking garage. The environment consists of a non-convex polyhedral outer boundary together with cylindrical and spherical internal obstacles.}
    \label{fig:grage_map}
\end{figure}

Realistic robotic environments are often inherently non-convex and populated by obstacles whose geometry cannot be adequately captured by spherical models. Examples include indoor environments with structural elements such as columns, beams, and walls, as well as parking-garage-like structures containing cylindrical and polyhedral features, as illustrated in Fig.~\ref{fig:grage_map}. These environments motivate the need for navigation-function constructions that operate directly in 3-D free spaces with complex obstacle geometry, without relying on global coordinate transformations.

\textbf{Related literature.}
Navigation functions are discussed in standard robotics texts such \linebreak as~\cite{choset2005principles}[Chap.~4], \cite{lavalle2006planning}[Chap.~8], and~\cite{lynch2017modern}[Chap.~10]. Extensions of navigation functions to partially known environments and sensor-driven settings were proposed in~\cite{filippidis2011adjustable,lionis2007locally,arslan2019sensor,vasilopoulos2022reactive,loizou2022mobile}. In addition, navigation-function have been used in the construction of control barrier functions~\cite{zehfroosh2022control} and in environments with moving obstacles~\cite{chen2020navigation,wei2023navigation}.

Beyond spherical obstacles, several works have explored navigation functions defined over more general geometries. Filippidis~\cite{filippidis2013navigation} studied conditions under which general smooth closed surfaces may serve as admissible obstacles, while Nicu~\cite{nicu2024design} investigated navigation functions over convex polyhedral obstacles. Despite these advances, the construction and analysis of navigation functions in 3-D environments containing non-spherical obstacles, particularly cylindrical obstacles and their intersections, remain relatively underexplored.

\textbf{Scope and contributions.}
This technical report focuses on the theoretical foundations of navigation functions in 3-D workspaces populated by spherical and cylindrical obstacles. We present a unified polynomial framework for encoding such obstacles, including pairwise intersecting configurations, and analyze the resulting navigation functions from a differential and topological perspective. In particular, the contributions of this report are:
\begin{itemize}
\item a polynomial encoding of spherical and cylindrical obstacles suitable for 3-D navigation functions;
\item explicit constructions of smooth navigation functions defined directly in 3-D free spaces;
\item analysis of critical points via gradient and Hessian structure, establishing polarity and admissibility under mild conditions;
\item treatment of pairwise intersecting obstacles using smooth composition techniques.
\end{itemize}

The emphasis of this report is on theoretical development and analysis. Algorithmic implementations, environment decomposition strategies, and large-scale planning architectures are intentionally excluded and will be addressed in future work.

The remainder of the report is organized as follows. Section~\ref{sec:workspace} introduces The 3-D model workspace and reviews the construction of navigation functions Section~\ref{sec:unique_min_proof} analyzes the critical point structure of the proposed navigation functions and establishes conditions under which the target is the unique minimum, addressing both disjoint and intersecting obstacles. Section~\ref{sec:simulations} presents simulation results in representative convex 3-D environments, including workspaces with composite obstacle geometries. Section~\ref{sec:topology} discusses the topological representation of the base obstacles underlying the proposed constructions. Next, Section~\ref{sec:spherical_robot} extends the framework to spherical robots via a transformation to a point-mass-robot model and reports corresponding simulation results. Lastly, conclusions are presented in Section~\ref{sec:conclutions}.

%%%%%%%%%%%%%%%%%%%%%%%%%%%%%%%%%%%%%%%%%%
%%%%%%%%%%%%%%%%%%%%%%%%%%%%%%%%%%%%%%%%%%
\section{3-D Model Workspaces}
\label{sec:workspace}
%%%%%%%%%%%%%%%%%%%%%%%%%%%%%%%%%%%%%%%%%%
%%%%%%%%%%%%%%%%%%%%%%%%%%%%%%%%%%%%%%%%%%
In our motion planning problem, the robot operates in a 3-D model \textit{workspace}, meaning the point robot workspace is represented as an $n$-dimensional ball of radius $r_0$, centered at the origin of $E^n$:
\begin{equation*}
    \mathcal{W} \triangleq \{\vec{x} \in E^n : \|\vec{x}\|^2 \leq r_0^2\} \quad n=2,3.
\end{equation*}
The focus of this work is on the case $n=3$.

Obstacles within the workspace are defined as either \textit{spherical} or \textit{cylindrical} regions. A \textit{spherical obstacle} is represented as an open ball of radius $r_j$, centered at $p_j \in \mathcal{W}$. A \textit{cylindrical obstacle} has radius $r_k$ and axis aligned with a unit direction $\hat{v}_k \in \mathbb{R}^3$, passing through a point $p_k \in \mathcal{W}$:
\begin{gather*}
    \mathcal{O}_s = \{x \in \mathbb{R}^3 : \|x - p_j\|^2 \leq r_j^2, \quad j = 1, \dots, m_s\}, \\
    \mathcal{O}_c = \{x \in \mathbb{R}^3 : \|\hat{v}_k \times (x - p_k)\|^2 \leq r_k^2, \quad k = 1, \dots, m_c\}, \\
    \mathcal{O} = \{\mathcal{O}_{s,1},\ldots, \mathcal{O}_{s,m_s}\} \cup \{\mathcal{O}_{c,1},\ldots, \mathcal{O}_{c,m_c}\} \\
    m = m_s + m_c.
\end{gather*}
Note that more complex obstacles can be formed by unions of spherical and cylindrical obstacles.

Thus, the \textit{free space} is defined as the 3-D region in $\mathbb{R}^3$:
\begin{equation*}
    \mathcal{F} \triangleq \mathcal{W} - \bigcup_{i=1}^{M} \text{int} (\mathcal{O}_i).
\end{equation*}

\noindent were $\text{int} (\mathcal{O}_i)$ denotes the interior set of the obstacles.

\begin{figure}[t!]
    \centering
    \includegraphics[width=0.7\linewidth]{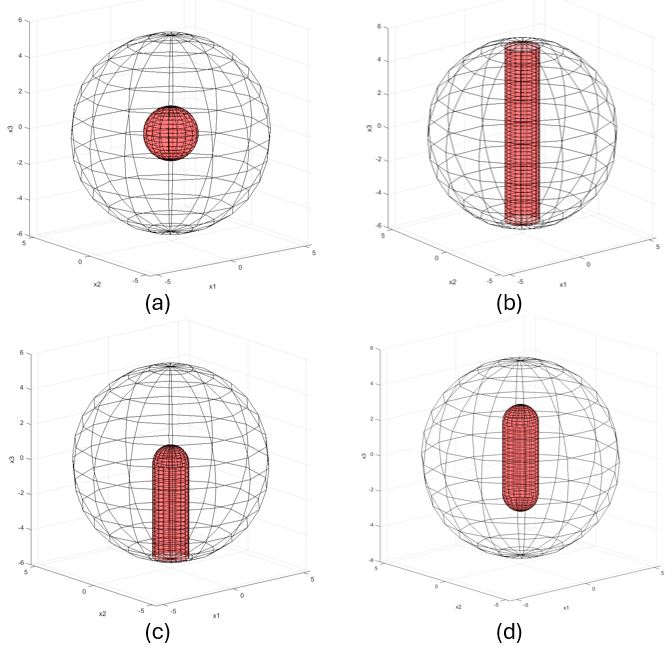}
    \caption{Different types of obstacles in a spherical room: (a) Spherical internal obstacle. (b) Full cylinder. (c) Half cylinder capped by a hemisphere. (d) Finite cylinder capped by two hemispheres.}
    \label{fig:obstacles}
\end{figure}

In this study, we consider three main types of obstacles:  
(i) \textit{spherical outer room walls},  
(ii) \textit{spherical obstacle} (Fig.~\ref{fig:obstacles}(a)),  
(iii) \textit{capped cylindrical obstacle}, which consist of a finite cylinder, capped at both ends by hemispheres (Fig.~\ref{fig:obstacles}(d)).  

\begin{figure}
    \centering
    \includegraphics[width=\linewidth]{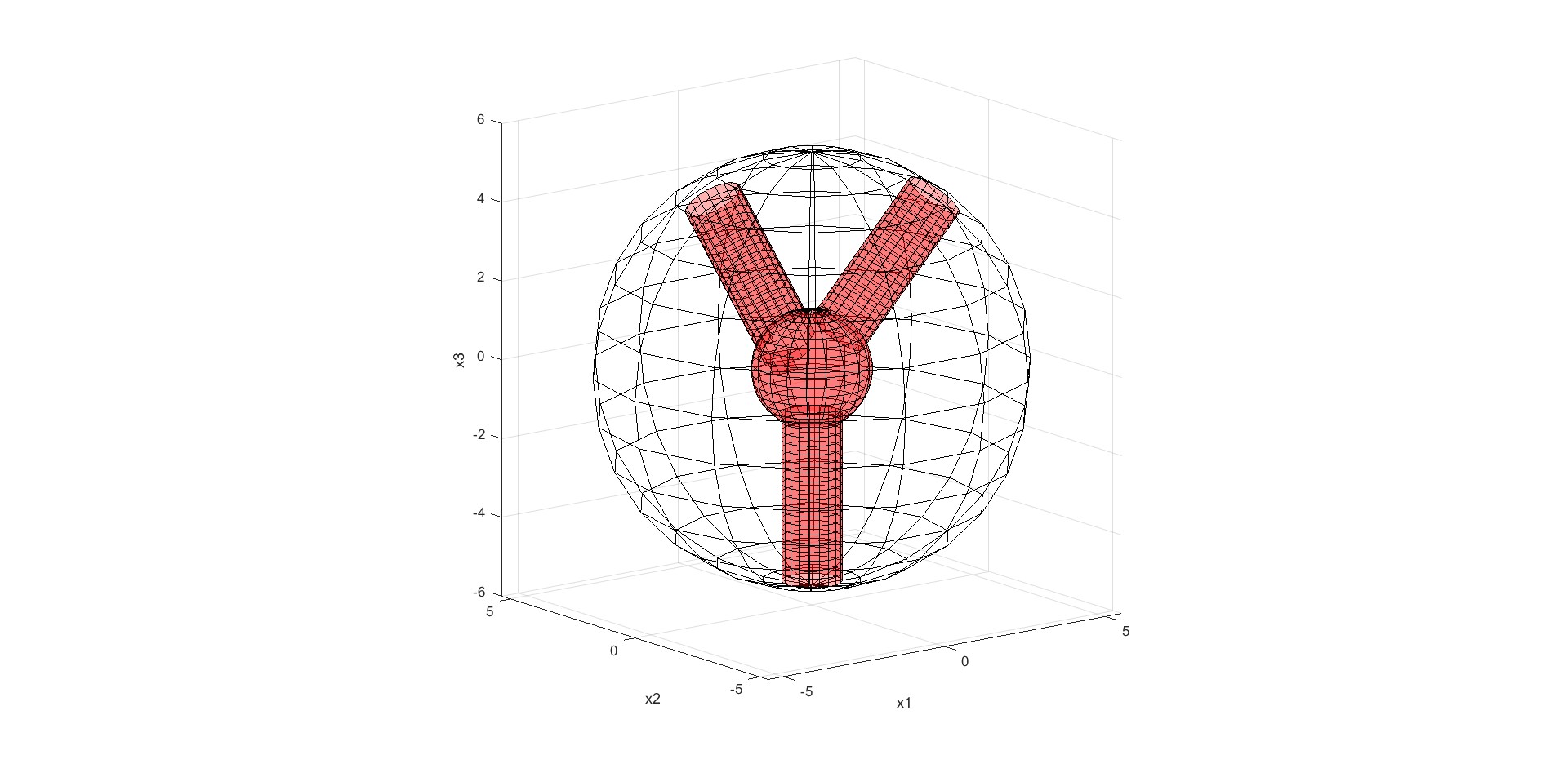}
    \caption{Composite obstacle consisting of pairwise unions between a spherical obstacle and three half-cylinder obstacles. The black shell represents the outer walls of the spherical room.}
    \label{fig:composite_obst}
\end{figure}

The capped cylindrical obstacles can appear in three forms:  
(i) \textit{full-cylinder obstacle}, whose hemispherical caps lie outside of the workspace, causing the cylinder to intersect the room outer wall at two distinct regions (two anchors), as illustrated in Fig.~\ref{fig:obstacles}(b);  
(ii) \textit{half-cylinder obstacle}, where only one hemispherical cap lies outside the workspace, resulting in a single intersection with the room’s outer wall (single anchor), as depicted in Fig.~\ref{fig:obstacles}(c);  
(iii) \textit{finite-cylinder obstacle}, where both hemispherical caps lie inside the workspace as shown in Fig.~\ref{fig:obstacles}(d).  
All other obstacles are formed by unions of these basic obstacle types. An example of a composite obstacle, created by merging a spherical obstacle with three half-cylinder obstacles, is shown in Fig.~\ref{fig:composite_obst}. Note that the black shell in Figs.~\ref{fig:obstacles} and~\ref{fig:composite_obst} represents the outer walls of the spherical room. The spherical walls of a room with radius $r_0$ are encoded by the following quadratic equation
\begin{equation}
    \beta_0(x) = {r_0}^2 - {\|x\|}^2 \quad x\in\mathbb{R}^3
\end{equation}

\noindent Similarly, spherical obstacles are encoded as
\begin{equation}
    \beta_{i,s}(x) = \|x - p_i\|^2 - r_i^2 \quad x\in\mathbb{R}^3
\end{equation}

\noindent where $r_i$ is the sphere radius and $p_i$ is the sphere center (Fig.~\ref{fig:obstacles}(a)). A full-cylindrical obstacle with two anchors (Fig.~\ref{fig:obstacles}(b)), encoded equation is defined as
\begin{equation}
    \beta_{i,c} = \|\hat{v}_i \times (x - p_i)\|^2 - r_i^2
\end{equation}

\noindent where $\hat{v}_i$ is the unit direction vector of the main axis of the cylinder, $p_i$ is a point along the axis and $r_i$ is the radius of the cylinder (Fig.~\ref{fig:obstacles}(c)). Thus, we can encode a general capped cylinder as
\begin{equation}
    \beta_{i,cc}(x) = 
    \begin{cases} 
       \|\hat{v}_i \times (x - {p}_{i,1})\|^2 - {r_i}^2 & \text{if } \hat{v}_i^T (x - {p}_{i,1}) \cdot \hat{v}_i^T (x - {p}_{i,2}) < 0 \\
        \|(x - {p}_{i,1})\|^2-{r_i}^2 & \text{if } \hat{v}_i^T(x - {p}_{i,1}) \leq 0 \text{ and } \hat{v}_i^T(x - {p}_{i,2}) \leq 0 \\
        \|(x - {p}_{i,2})\|^2-{r_i}^2 & \text{if } \hat{v}_i^T(x - {p}_{i,1}) \geq 0 \text{ and } \hat{v}_i^T(x - {p}_{i,2}) \geq 0 \\
    \end{cases}
    \label{eq:beta_cc}
\end{equation}

\noindent Here, $p_{i,1}$ and $p_{i,2}$ denote the endpoints of the cylinder, $\hat{v}_i$ represents the cylinder axis direction, defined as $\hat{v}_i = \widehat{p_{i,2} - p_{i,1}}$, and $r_i$ is the radius of both the cylinder and its hemispherical caps. The definition in Eq.~(\ref{eq:beta_cc}) ensures that $\beta_{i,cc}$ forms a continuously differentiable function ($C^1$). Note that a sphere can be defined as a capped-cylinder where $p_{i,1} = p_{i,2} = p_i$. However, for simplicity, we will treat them as separate obstacles. 

%%%%%%%%%%%%%%%%%%%%%%%%%%%%%%%%%%%%%%%%
\subsection{Navigation Function Construction}
\label{sec:nav_fun}
%%%%%%%%%%%%%%%%%%%%%%%%%%%%%%%%%%%%%%%%
A \textit{navigation function} is a scalar-valued function, $\hat{\varphi}:\mathcal{F}\rightarrow[0,1]$, that satisfies the following properties within a given free space:

\begin{enumerate}
\setlength{\itemsep}{1pt} 
    \item It is smooth on an open set of $\mathbb{R}^n$ that contains $\mathcal{F}$, where $n=3$ in our case.
    \item It has a unique global minimum at the target, $p_d$, such that $\hat{\varphi}(p_d)=0$.
    \item All its saddle points are non degenerate critical points in $\mathcal{F}$.
    \item It is admissible, meaning that it is bounded and is uniformly maximal with value of 1 at the boundary of $\mathcal{F}$.
\end{enumerate}

By leveraging the gradient of a navigation function, $\nabla\hat{\varphi}(x)$, a force-feedback controller can be designed to steer an omnidirectional point mass robot from almost any initial state to the target location. In this study, we consider two navigation functions, $\varphi$ and $\psi$, both derived from a base navigation function, $\hat{\varphi}$.

%%%%%%%%%%%%%%%%%%%%%%%%%%%%%%%%%%%%%%%%
\subsection{Base Navigation Function Construction}
\label{sec:base_nav_fun}
%%%%%%%%%%%%%%%%%%%%%%%%%%%%%%%%%%%%%%%%

We define the \textit{base navigation function} as:
\begin{equation}
    \hat{\varphi}({x}) = \frac{{\gamma_d}^k(x)}{\beta(x)} \quad k\in \mathbb{N}
    \label{eq:phi_hat}
\end{equation}
where $k$ is the tuning parameter and $\gamma_d$ represents the attractive potential toward the target
\begin{equation}
    \gamma_d(x) = \|x - {p}_d\|^2. 
    \label{eq:gamma}
\end{equation}
where $x$ and ${p}_d$ are the current and target locations. The denominator term $\beta(x)$ accounts for the repulsive effects from the boundaries of the free space $\mathcal{F}$ and is expressed as the product of the individual obstacle repulsion terms
\begin{equation}
    \beta(x) = \prod_{i=0}^{m} \beta_i(x) = \beta_0(x) \cdot \prod_{i=1}^{m}\beta_i(x) 
    \label{eq:beta}
\end{equation}
\noindent where $\beta_0(x)$ encodes the repulsive effect for the room walls and $\beta_i(x)$ for $i=1\dots m$ encodes the repulsive effect for the $i$-th obstacle.

%%%%%%%%%%%%%%%%%%%%%%%%%%%%%%%%%%%%%%%%%%%%%%%
\subsection{Base Function Gradient and Hessian}
%%%%%%%%%%%%%%%%%%%%%%%%%%%%%%%%%%%%%%%%%%%%%%%

To determine the conditions under which $\mathcal{F}$ has a unique global minimum at $p_d$, we first compute the gradient and Hessian of $\gamma_d(x)$ and $\beta(x)$:

\begin{equation}
    \nabla \gamma_d(x) = 2 ({x} - {p_d})
    \label{eq:grad_gamma}
\end{equation}
\begin{equation}
    D^2 \gamma_d(x) = 2I
    \label{eq:hessian_gamma}
\end{equation}

Since $\beta(x)$ is defined as the product of individual terms $\beta_i(x)$ (as shown in Eq.~(\ref{eq:beta})), its gradient is the sum
\begin{equation}
    \nabla \beta(x) = \sum_{i=0}^{m} \nabla \beta_i(x) \cdot \bar{\beta}_i(x)
    \label{eq:grad_beta}
\end{equation}
where $\bar{\beta}_i(x)$ is the omitted product
\begin{equation*}
    \bar{\beta}_i(x) = \prod_{\substack{j=0 \\ j\neq i}}^{M} \beta_j(x).
\end{equation*}

\noindent The gradients of the individual terms take the form

\begin{gather}
    \nabla \beta_0(x) = -2{x} \label{eq:grad_b_0} \\
    \nabla \beta_{i,s}(x) = 2\left({x} - {p_i}\right)  \label{eq:grad_b_s} \\
    \nabla\beta_{i,cc}(x) = 
    \begin{cases} 
       2\big(({x} - {p}_{i,1}) - (\hat{v}_i \cdot ({x} - {p}_{i,1})) \hat{v}_i \big) & \text{if } \hat{v}_i^T ({x} - {p}_{i,1}) \cdot \hat{v}_i^T ({x} - {p}_{i,2}) < 0 \\
        2\left({x} - {p}_{i,1}\right) & \text{if } \hat{v}_i^T({x} - {p}_{i,1}) \leq 0 \text{ and } \hat{v}_i^T({x} - {p}_{i,2}) \leq 0 \\
        2\left({x} - {p}_{i,2}\right)  & \text{if } \hat{v}_i^T({x} - {p}_{i,1}) \geq 0 \text{ and } \hat{v}_i^T({x} - {p}_{i,2}) \geq 0 
    \end{cases} \label{eq:grad_b_cc}
\end{gather}

For half-cylinder and finite-cylinder obstacles, at the gluing circle between the cylinder and hemispherical caps associated with the endpoint $p_{i,j}$, the condition $\hat{v}_i \perp ({x} - {p}_{i,j})$ holds. Consequently, the gradients satisfy $\nabla \beta_{i,s}(x) = \nabla \beta_{i,c}(x)$, ensuring that $\beta_{i,cc}(x)$ is continuously differentiable ($C^1$).  

Using the expressions (\ref{eq:grad_b_0}), (\ref{eq:grad_b_s}) and (\ref{eq:grad_b_cc}), the Hessian of each $\beta_i$ is given by

\begin{gather}
    D^2 \beta_0(x) = -2I \\
    D^2 \beta_{i,s}(x) = 2I \\
    D^2\beta_{i,cc}(x) = 
    \begin{cases} 
      2\left[I - \hat{v}_i \cdot \hat{v}_i^T \right] & \text{if } \hat{v}_i^T ({x} - {p}_{i,}) \cdot \hat{v}_i^T ({x} - {p}_{i,2}) < 0 \\
        2I & \text{if } \hat{v}_i^T ({x} - {p}_{i,1}) \cdot \hat{v}_i^T ({x} - {p}_{i,2}) \geq 0\\
    \end{cases}
\end{gather}

Notably, if one defines $\hat{v}_i = \vec{0}$ for spherical obstacles, the gradient and Hessian derived for cylindrical obstacles can also be applied to spherical obstacles. However, for clarity we distinguish between spherical and cylindrical obstacles.

%%%%%%%%%%%%%%%%%%%%%%%%%%%%%%%%%%%%
\subsection{Composite Navigation Functions}
\label{sec:composite_vav_func}
%%%%%%%%%%%%%%%%%%%%%%%%%%%%%%%%%%%%
This study evaluates two composite navigation functions, $\varphi$ and $\psi$. 
Both functions are constructed using the same three steps, applied in different orders. To construct these functions we first define a \textit{normalizing function} $\sigma: \mathbb{R}^+\rightarrow[0,1]$

\begin{equation}
    \sigma(x) = \frac{x}{1 + x} \quad\quad x\in\mathbb{R}^+
\end{equation}

\noindent Additionally, we employ a \textit{sharpening function} $\sigma_d: \mathbb{R}^+\rightarrow\mathbb{R}^+$

\begin{equation}
    \sigma_d(x) = x^{\frac{1}{k}}  \quad\quad x\in\mathbb{R}^+ \quad k\in \mathbb{N}
\end{equation}

\noindent where $k$ is the tuning parameter used in Eq.~(\ref{eq:phi_hat}). This function ensures that the critical point at the target remains non-degenerate. Let us define the navigation function $\varphi$ to be
\begin{equation}
    \varphi({x}) = \sigma_d \circ \sigma \circ \hat{\varphi}({x}) = \frac{\gamma_d(x)}{\left({\gamma_d(x)}^k + \beta(x)\right)^{1/k}} 
\end{equation}

\noindent Thus, the gradient of $\varphi$ is given by

\begin{equation}
    \nabla\varphi({x}) = \frac{1}{d(x)^2}\big(2d(x)\cdot({x}-{p_d})-\gamma_d(x)\cdot\nabla d(x)\big) \\
\end{equation}
\noindent where
\begin{equation*}
     d(x) = \left({\gamma_d(x)}^k + \beta(x)\right)^{1/k}
\end{equation*}
\noindent and
\begin{equation*}
    \nabla d(x) = \frac{1}{k} \cdot  \left({\gamma_d(x)}^k + \beta(x)\right)^\frac{1-k}{k} \cdot \left(2k \cdot {\gamma_d(x)}^{k-1} \cdot ({x}-{p_d}) + \sum_{i=0}^{m}  \nabla\beta_i(x) \cdot \prod_{\substack{j=0 \\ j\neq i}}^{m} \beta_j(x)\right)
\end{equation*}

Similarly, let us define the alternate navigation function $\psi$ to be
\begin{equation}
    \psi({x}) = \sigma \circ \sigma_d \circ \hat{\varphi}({x})  = \frac{\gamma_d(x)}{\gamma_d(x) + \beta(x)^{1/k}}
\end{equation}

\noindent Thus, the gradient of $\psi$ is given by

\begin{equation}
    \nabla\psi({x}) = \frac{1}{d(x)^2} \big(2d(x)\cdot({x}-{p_d})-\gamma_d(x)\cdot\nabla d(x)\big)
    \label{eq:grad_psi}
\end{equation}
\noindent where
\begin{equation*}
    d(x) = {\gamma_d(x)} + \beta(x)^{1/k}
\end{equation*}
\begin{equation*}
    \nabla d(x) = 2({x}-{p_d}) +\frac{1}{k} \cdot \beta(x)^\frac{1-k}{k} \cdot \left( \sum_{i=0}^{m}  \nabla\beta_i(x) \cdot \prod_{\substack{j=0 \\ j\neq i}}^{m} \beta_j(x)\right)
\end{equation*}

%%%%%%%%%%%%%%%%%%%%%%%%%%%%%%%%%%%%%%%%%%%%%%%%%
%%%%%%%%%%%%%%%%%%%%%%%%%%%%%%%%%%%%%%%%%%%%%%%%%
\section{Ensuring a Unique Minimum at the Target}
\label{sec:unique_min_proof}
%%%%%%%%%%%%%%%%%%%%%%%%%%%%%%%%%%%%%%%%%%%%%%%%%
%%%%%%%%%%%%%%%%%%%%%%%%%%%%%%%%%%%%%%%%%%%%%%%%%

The proof for the existence of a unique minimum in a 2-D setting with disjoint disc obstacles appears in ~\cite{rimon1988exact}. However, this proof does not directly extend to the 3-D case due to the presence of non-spherical obstacles, such as cylinders. With slight modifications, the approach can be adapted for a 3-D setting. This section provides the necessary proof in the 3-D setting.

Let $\epsilon > 0$ define an open neighborhood around the boundary of the free space $\mathcal{F}$, such that $\mathcal{B}_i(\epsilon)=\{\vec{x}\in E^3 : 0 < \beta_i < \epsilon\}$. Using this definition, we can partition the free space into five distinct subsets:
\begin{enumerate}
    \item The destination point
        \begin{equation*}
            \{p_d\};
        \end{equation*}
    \item The boundary of the free space
        \begin{equation*}
            \partial\mathcal{F} \triangleq \beta^{-1}(0);
        \end{equation*}
    \item The region near the internal obstacles
        \begin{equation*}
            \mathcal{F}_0(\epsilon) \triangleq \bigcup_{i=1}^{M} \mathcal{B}_i(\epsilon) -  \{p_d\};
        \end{equation*}
    \item The region near the workspace outer boundary
        \begin{equation*}
            \mathcal{F}_1(\epsilon) \triangleq \mathcal{B}_0(\epsilon) - \left(\{p_d\} \cup \mathcal{F}_0(\epsilon)\right);
        \end{equation*}
    \item The region away from the obstacles
        \begin{equation*}
            \mathcal{F}_2(\epsilon) \triangleq \mathcal{F} - \left(\{p_d\} \cup \partial\mathcal{F} \cup \mathcal{F}_0(\epsilon) \cup \mathcal{F}_1(\epsilon)\right);
        \end{equation*}
\end{enumerate}

\begin{figure}[t!]
    \centering
    \includegraphics[width=0.7\linewidth]{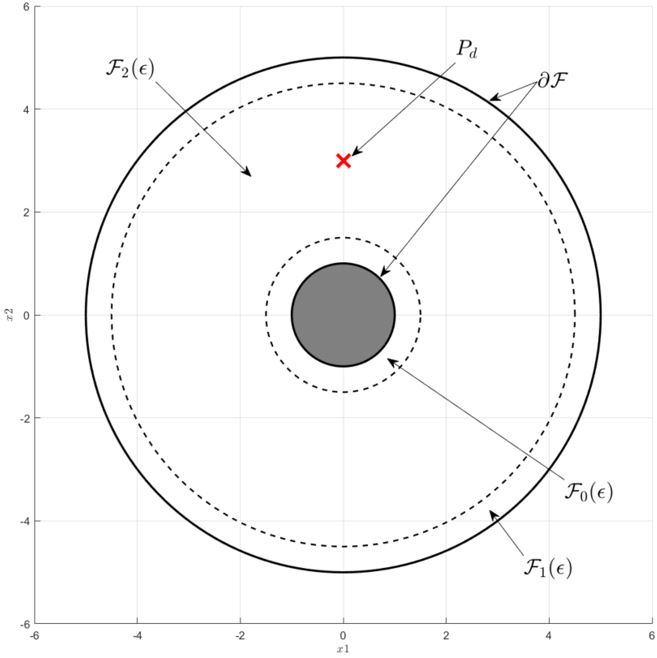}
    \caption{2-D illustration of the five regions into which the free space is divided}
    \label{fig:regions}
\end{figure}

An illustration of these regions is shown in Fig.~\ref{fig:regions}. We assume that $\epsilon$ is sufficiently small so that the exclusion of $p_d$ from both $\mathcal{F}_0(\epsilon)$ and $\mathcal{F}_1(\epsilon)$ is redundant, and that $\mathcal{F}_0(\epsilon) \subseteq \mathcal{F}$.

Our polynomial navigation functions can be expressed as the ratio of two terms, thus, let us examine the gradient and Hessian of the general ratio function
\begin{equation*}
    \rho \triangleq \frac{\nu}{\delta}
\end{equation*}
At a critical point, the gradient of this function satisfies $\nabla\rho = \vec{0}$. To determine the nature of the critical point, the Hessian $D^2\rho$ is required. Specifically, a critical point is classified as a local minimum if all eigenvalues of the Hessian $D^2\rho$ are positive. The gradient of such a navigation function can be written as:
\begin{equation}
     \nabla\rho = \frac{1}{\delta^2}\left(\delta\nabla\nu - \nu\nabla\delta\right)
     \label{eq:grad_phi}
\end{equation}
Thus, the Hessian of $\rho$ is given by:
\begin{equation}
    D^2\rho = \frac{1}{\delta^2}[\delta D^2\nu + \nabla\nu\nabla\delta^T - \nabla\delta\nabla\nu^T - \nu D^2\delta] + \delta^2\nabla\rho\left(\nabla\frac{1}{\delta^2}\right)^T
\end{equation}

Since at the critical point $\nabla\rho = \vec{0}$, it follows from Eq.~(\ref{eq:grad_phi}) that $\nabla\nu = \rho \nabla\delta$. Therefore, the Hessian at a critical point simplifies to:
\begin{equation}
    D^2\rho|\mathcal{C}_p = \frac{1}{\delta^2}[\delta D^2\nu - \nu D^2\delta]
    \label{eq:hessian}
\end{equation}

Now, let us show that our functions are valid navigation functions by evaluating the critical points. As demonstrated in \cite{Rimon1988}, composing the navigation function $\hat{\varphi}$ with a monotonic function does not alter the location or type of critical points. Consequently, substituting different navigation functions in various parts of this proof remains valid.

%%%%%%%%%%%%%%%%%%%%%%%%%%%%%%%%%%%%%%%%
\subsection{Destination and Boundary of \texorpdfstring{$\mathcal{F}$}{F}}
%%%%%%%%%%%%%%%%%%%%%%%%%%%%%%%%%%%%%%%%
In this section, we demonstrate that $\varphi$ attains a global minimum at the target. Additionally, we analyze the critical points of $\varphi$ along the boundary of the free space, $\partial\mathcal{F}$.

\begin{claim}
    For a valid workspace $\mathcal{W}$, the destination point, $p_d$ is a non-degenerate local minimum of $\varphi$.
\end{claim}

\begin{proof}
    Evaluating $\varphi$ at $p_d$ using $\nabla\varphi$ and $D^2\varphi$ confirms the existence of a non-degenerate local minimum at the destination.
\end{proof}

\begin{figure}
    \centering
    \includegraphics[width=\linewidth]{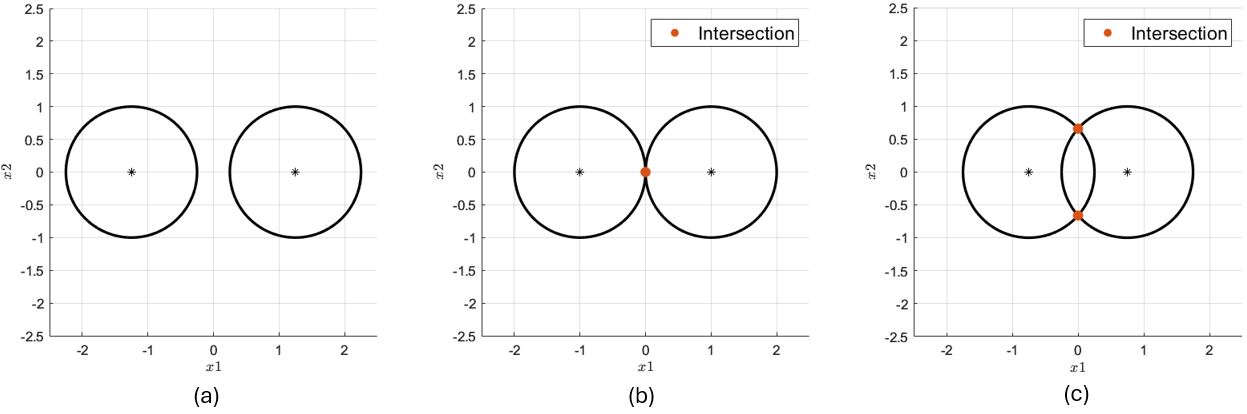}
    \caption{Spatial relationships between two 2-D disc obstacles. (a) Disjoint obstacles. (b) Tangent obstacles with a single shared point. (c) Intersecting obstacles with two intersection points.}
    \label{fig:intersecting_obstacles}
\end{figure}

We can distinguish between three different spatial relationships between two obstacles: (i) disjoint obstacles, (ii) tangent obstacles, and (iii) intersecting obstacles. Disjoint obstacles, as illustrated in Fig.~\ref{fig:intersecting_obstacles}(a), do not share any common points. Tangent obstacles, shown in Fig.~\ref{fig:intersecting_obstacles}(b), touch at exactly one point. This is also true in a 3-D setting, except for parallel cylinders, where the tangent forms a line parallel to the main axis of the cylinders. Intersecting obstacles, illustrated in Fig.~\ref{fig:intersecting_obstacles}(c), share multiple intersection points. In 3D, the intersection between two obstacles forms a closed curve, except in the case of two parallel cylinders, where the intersection is represented by two parallel lines.

\begin{claim}
    For a valid workspace $\mathcal{W}$ with no tangent obstacles, all critical points of $\varphi$ are in the interior of $\mathcal{F}$.
\end{claim}
\begin{proof}
    The gradient of $\varphi$ at a point $p_0 \in \partial\mathcal{F}$, where $\beta_i(p_0) = 0$, is given by:
    \begin{align*}
        \nabla\varphi(p_0) &= \frac{1}{({\gamma_d}^k+\beta)^\frac{2}{k}}\left(({\gamma_d}^k+\beta)^\frac{1}{k}\nabla\gamma_d - \gamma_d\nabla({\gamma_d}^k+\beta)^\frac{1}{k} \right)|_{p_0}\\
        &= -\frac{1}{k}{\gamma_d}^{-k}\left(\prod_{\substack{j=0 \\ j\neq i}}^{M}\beta_j \right)\nabla\beta_i.
    \end{align*}
    
    It is evident that for disjoint obstacles, $\nabla\varphi(p_0) \neq 0$. 
    
    However, in the case of two tangent or intersecting obstacles, where $\beta_i(p_0) = \beta_k(p_0) = 0$, the gradient along the intersection curve satisfies $\nabla\varphi(p_0) = \vec{0}$, causing these critical points to appear as a curve of local maxima. This issue can be addressed by merging both obstacles using the \textit{p}-Rvachev function~\cite{shapiro1999implicit}:
    \begin{equation*}
        \beta_{ik} = R_p(\beta_i,\beta_k) = \beta_i + \beta_k - \left(\beta_i^p + \beta_k^p \right)^{\frac{1}{p}} \quad p>1
    \end{equation*}
    
    \begin{figure}[b!]
        \centering
        \includegraphics[width=\linewidth]{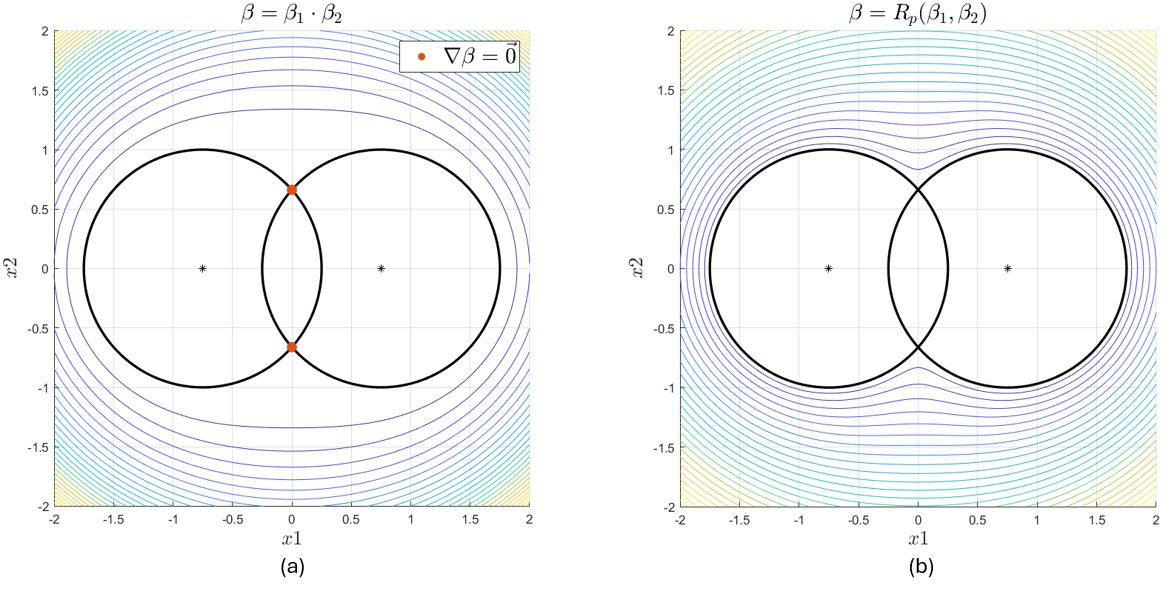}
        \caption{$\beta_{ik}$ value for intersecting 2-D obstacles. (a) $\beta_{ik}=\beta_i\cdot\beta_k$. (b) $ \beta_{ik} = R_p(\beta_i,\beta_k)$}
        \label{fig:beta_val}
    \end{figure}
    
    The \textit{p}-Rvachev function is a smooth function in the interior of the workspace and ensures that $\beta_{ik}=0$ on the boundary of the obstacles. An example for both separate and merged obstacles are shown in Fig.~\ref{fig:beta_val}(a) and Fig.~\ref{fig:beta_val}(b). Accordingly, the gradient of the combined obstacle is given by
    \begin{equation*}
        \nabla\beta_{ik}= \nabla\beta_i + \nabla\beta_k - \left(\beta_i^p + \beta_k^p\right)^{\frac{1-p}{p}}\left( \beta_i^{p-1}\nabla{\beta_i} + \beta_k^{p-1}\nabla{\beta_k}  \right)
    \end{equation*}

    \noindent Notice that the gradient on the boundary of the merged obstacles is given by
    \begin{equation}
        \nabla\beta_{ik}=
        \begin{cases}
            \nabla{\beta_i}, & \text{if } \beta_i=0 \text{ and } \beta_k\neq0 \\
            \nabla{\beta_k}, & \text{if } \beta_k=0 \text{ and } \beta_i\neq0\\
            \nabla{\beta_i}+\nabla{\beta_k}, & \text{if } \beta_i=\beta_k=0 \\
        \end{cases}
    \end{equation}

\noindent Since the direction of both $\nabla\beta_i$ and $\nabla\beta_k$ points away from the center of their respective obstacle, it follows that at the intersection of two obstacles, $\nabla\beta_i$ and $ \nabla\beta_k $ are not parallel ($ \nabla\beta_i \nparallel \nabla\beta_k $). Consequently, $ \nabla\beta_{ik} \neq \vec{0} $.

\end{proof}
However, note that for tangent obstacles, $\nabla\beta_{ik}(p_0)$ points into the obstacle region and becomes zero when the obstacles have the same radius.

%%%%%%%%%%%%%%%%%%%%%%%%%%%%%%%%%%%%%%%%%%%%%%%%%%%%%%%%%%
\subsection{Interior of \texorpdfstring{$\mathcal{F}$}{F} for Disjoint Obstacles}
\label{sec:interior_F}
%%%%%%%%%%%%%%%%%%%%%%%%%%%%%%%%%%%%%%%%%%%%%%%%%%%%%%%%%%
In this section, we demonstrate that for a disjoint obstacle workspace $\mathcal{W}$, the only local minimum within the interior of $\mathcal{F}$ is at the destination. To achieve this, we utilize $\hat{\varphi}$. First, we establish that apart from the target, all other critical points can be arbitrarily shifted closer to the boundary of the free space by imposing a constraint on the tuning parameter $k$. Subsequently, we show that none of these points can be local minima by analyzing the quadratic form of the Hessian at the critical points.

\begin{claim}
    For every $\epsilon > 0$, there exists an $N(\epsilon) \in \mathbb{N}$ such that if $k \geq N(\epsilon)$, then $\hat{\varphi}$ has no critical points in $\mathcal{F}_2(\epsilon)$, the interior of $\mathcal{F}$ away from the obstacles.
\end{claim}

\begin{proof}
    Using Eq.~(\ref{eq:grad_phi}), we obtain that at critical points of $\hat{\varphi}$:
    \begin{equation}
        k\beta\nabla\gamma_d = \gamma_d\nabla\beta.
        \label{eq:grad_at_Cp}
    \end{equation}
    
    Since $\|\nabla\gamma_d\| = 2\sqrt{\gamma_d}$, as can be verified by comparing Eq.~(\ref{eq:gamma}) and Eq.~(\ref{eq:grad_gamma}), taking the magnitude of both sides gives
    \begin{equation*}
        2k\beta = \sqrt{\gamma_d} \|\nabla\beta\|.
    \end{equation*}
    
    A sufficient condition for this equality \textit{not} to hold for all $x \in \mathcal{F}_2(\epsilon)$ is
    \begin{equation*}
        k > \frac{1}{2} \frac{\sqrt{\gamma_d} \|\nabla\beta\|}{\beta}
    \end{equation*}
    
    For a given $\epsilon$, using Eq.~(\ref{eq:grad_beta}), the upper bound of the right-hand side is given by
    \begin{equation}
        \begin{aligned}
          \frac{1}{2} \frac{\sqrt{\gamma_d} \|\nabla\beta\|}{\beta} 
          \leq \frac{1}{2} \sqrt{\gamma_d} \sum_{i=0}^{M} \frac{\bar{\beta_i}}{\beta} \|\nabla\beta_i\| \\
          < \frac{1}{2} \frac{1}{\epsilon} \max_{\mathcal{W}}\{\sqrt{\gamma_d}\} 
          \sum_{i=0}^{M} \max_{\mathcal{W}}\{\|\nabla\beta_i\|\} \triangleq N(\epsilon)
        \end{aligned}
        \label{eq:N_eps}
    \end{equation}
    since $\beta_j \geq \epsilon$ for all $j \in \{0, \dots, M\}$ in the region $\mathcal{F}_2(\epsilon)$.
\end{proof}

If $p_c \in \mathcal{F}_0(\epsilon) \cap \mathcal{C}_{\hat{\varphi}}$, then $p_c$ belongs to at least one of the $\mathcal{B}_i(\epsilon)$ for some $i \in \{1, \dots, M\}$, meaning it is $\epsilon$-close to the $i$-th obstacle. By evaluating the quadratic form along a unit vector orthogonal to $\nabla \beta_i$, we can demonstrate that ${D}^2\hat{\varphi}(p_c)$ has at least one negative eigenvalue, implying that $p_c$ is not a local minimum.

\begin{claim}
    For any valid workspace, with disjoint obstacles, there exists $\epsilon_0 > 0$ such that $\hat{\varphi}$ has no local minima in $\mathcal{F}_0(\epsilon)$, the region near the internal obstacles, for all $\epsilon < \epsilon_0$.
\end{claim}

\begin{proof}
    Using Eq.~(\ref{eq:hessian}), we derive the following expression for the Hessian of $\hat{\varphi}$ at $p_c$:
    \begin{equation}
        \begin{aligned}
            {D}^2\hat{\varphi}(p_c) &= \frac{1}{\beta^2} \big[ \beta D^2 ({\gamma_d}^k) - {\gamma_d}^k D^2 \beta \big] \\
            &= \frac{{\gamma_d}^{k-2}}{\beta^2} \left( k\beta \big[ \gamma_d D^2{\gamma_d} + (k-1) \nabla{\gamma_d} \nabla{\gamma_d}^T \big] - {\gamma_d}^2 D^2{\beta} \right)
        \end{aligned}
        \label{eq:hessian_at_Cp}
    \end{equation}
    
    Taking the outer product of Eq.~(\ref{eq:grad_at_Cp}), we obtain
    \begin{equation}
        (k\beta)^2 \nabla{\gamma_d} \nabla{\gamma_d}^T = {\gamma_d}^2 \nabla{\beta} \nabla{\beta}^T
        \label{eq:grad_gamma_product}
    \end{equation}
    
    Substituting for $\nabla{\gamma_d} \nabla{\gamma_d}^T$ in Eq.~(\ref{eq:hessian_at_Cp}) and using $\beta = \beta_i \bar{\beta_i}$, we get
    \begin{align*}
        {D}^2\hat{\varphi}(p_c) = \frac{{\gamma_d}^{k-1}}{\beta^2} &\Bigg( k\beta {D}^2\gamma_d + \left( 1-\frac{1}{k} \right) \frac{\gamma_d}{\beta} \Big[ {\beta_i}^2 \nabla{\bar{\beta_i}} \nabla{\bar{\beta_i}}^T \\
        &+ 2\beta_i \bar{\beta_i} (2\nabla{\bar{\beta_i}} \nabla{{\beta_i}}^T) + {\bar{\beta_i}}^2 \nabla{\beta_i} \nabla{\beta_i}^T \Big] \\
        &-\gamma_d \Big[ \beta_i D^2 \bar{\beta_i} + 2(\nabla \bar{\beta_i}^T  \nabla{\beta_i}) + \bar{\beta_i} D^2 {\beta_i} \Big] \Bigg)
    \end{align*}
    
    Now, we evaluate the quadratic form associated with ${D}^2\hat{\varphi}(p_c)$ at $\hat{u} = \widehat{\nabla \beta_i(p_c)\times \hat{v}_i}$. 
    
    Recalling that ${D}^2\gamma_d = 2I$ and ${D}^2\beta_i = 2I - 2\hat{v}_i \hat{v}_i^T$, the resulting quadratic form is
    \begin{equation}
    \begin{aligned}
         \frac{\beta^2}{{\gamma_d}^{k-1}} \hat{u}^T D^2\hat{\varphi}(p_c) \hat{u} &= 2k\beta - 2\gamma_d\bar{\beta_i} \\ 
        &+\hat{u}^T \Bigg[ \left( 1 - \frac{1}{k} \right) \frac{\gamma_d}{\beta} {\beta_i}^2 {\nabla\bar{\beta_i}} \nabla{\bar{\beta_i}}^T - \gamma_d \beta_i D^2\bar{\beta_i} \Bigg]\hat{u}
    \end{aligned}
    \label{eq:q_form}
    \end{equation}
    
    Considering the inner product of Eq.~(\ref{eq:grad_at_Cp}) with $\nabla{\gamma_d}$ we obtain
    \begin{equation}
        4k\beta = \nabla{\beta} \cdot \nabla{\gamma_d} = \bar{\beta_i} \nabla{\beta_i} \cdot \nabla{\gamma_d} + \beta_i \nabla{\bar{\beta_i}} \cdot \nabla{\gamma_d}
    \end{equation}
    
    Substituting for $2k\beta$ in Eq.~(\ref{eq:q_form}) results in the following equation
    \begin{equation}
    \begin{aligned}
         \frac{\beta^2}{{\gamma_d}^{k-1}} \hat{u}^T D^2\hat{\varphi}&(p_c) \hat{u} = 2\bar{\beta_i} \left( \frac{1}{4}\nabla{\beta_i} \cdot \nabla{\gamma_d} - \gamma_d \right) \\ 
        &+ \beta_i\left( \frac{1}{2} \nabla{\bar{\beta_i}} \cdot \nabla{\gamma_d} + \gamma_d \hat{u}^T \Bigg[ \left( 1 - \frac{1}{k} \right) \frac{1}{\bar{\beta_i}} {\nabla\bar{\beta_i}} \nabla{\bar{\beta_i}}^T - D^2\bar{\beta_i} \Bigg] \hat{u} \right)
    \end{aligned}
    \label{eq:q_form_final}
    \end{equation}
    
    Note that this equation was also derived in \cite{rimon1988exact}, indicating that it holds for both spherical and cylindrical obstacles. Where $\hat{u}=\widehat{\nabla{\beta_i(p_c)}^\perp}$ for spherical obstacles.

    Let us now define the following term:
    \begin{equation}
        Q_i = \frac{1}{4} \nabla\gamma_d \cdot \nabla\beta_i - \gamma_d
    \end{equation}
    
    It was shown in~\cite{Rimon1988} that the maximum value of $Q_i$ for a spherical obstacle is:
    \begin{equation}
        \max\limits_{\mathcal{B}_{i,sphere}(\epsilon)}\{Q_i\} = \|p_d - p_i\|\left(\sqrt{\epsilon + {r_i}^2} - \|p_d - p_i\|\right)
        \label{eq:ui_sphere}
    \end{equation}
    
    However, in our case, there are also cylindrical obstacles for which the value of $Q_i$ is:
    \begin{equation*}
        Q_i = \left( (x - p_d)^T (x - p_i) -  (x - p_d)^T (x - p_d)\right) - \hat{v}_i^T (x - p_i) \cdot \hat{v}_i^T (x - p_d) 
    \end{equation*}
    
    To find the upper bound for this expression, it is sufficient to determine the maximum of the left term and the minimum of the right term. This leads to the following result:
    \begin{equation}
        \begin{aligned}
            \max\limits_{\mathcal{B}_{i}(\epsilon)} \{Q_i\} = (p_d - p_i)^T &\left( \left(a(1-a)+b\right) \hat{v}_i + \left(\sqrt{\epsilon+{r_i}^2} - \sqrt{\beta_i(p_d)+{r_i}^2}\right) \widehat{\nabla\beta_i(p_d)} \right)\\
            &a = \frac{\sqrt{\epsilon + {r_i}^2}}{\sqrt{\beta_i(p_d) - {r_i}^2}} \\
            &b =
            \begin{cases}
                -2r_0, & \text{if } \hat{v}_i^T (p_d - p_i) > 0 \\
                0, & \text{if } \hat{v}_i^T (p_d - p_i) = 0 \\
                2r_0, & \text{if } \hat{v}_i^T (p_d - p_i) < 0
            \end{cases}
        \end{aligned}
    \end{equation}
    Since $p_i$ is an arbitrary point along the main axis of the cylinder, we select it such that $(p_d - p_i) \perp \hat{v}_i$. This choice allows us to simplify the maximum value of $Q_i$ to

    \begin{equation}
        \max\limits_{\mathcal{B}_{i}(\epsilon)} \{Q_i\} = (p_d - p_i)^T \left(\left(\sqrt{\epsilon+{r_i}^2} - \sqrt{\beta_i(p_d)+{r_i}^2}\right) \widehat{\nabla\beta_i(p_d)} \right)
        \label{eq:ui_cylinder}
    \end{equation}
    
    It is evident that for both spherical and cylindrical obstacles, Eqs.~(\ref{eq:ui_sphere}) and~(\ref{eq:ui_cylinder}) are identical. Thus, Eq.~(\ref{eq:ui_sphere}) can be used for both types of obstacles.
    
    Returning to Eq.~(\ref{eq:q_form_final}), the second term is proportional to $\beta_i$, meaning it can be made arbitrarily small by an appropriate choice of $\epsilon$. However, it may still remain positive. Therefore, to ensure that the quadratic form is negative, we must first guarantee that the first term, $2\bar{\beta_i}{Q_i}$, is negative. Using Eq.~(\ref{eq:ui_cylinder}), this condition holds when
  \begin{equation}
    \epsilon < \epsilon'_{0i} \triangleq \max \{ \epsilon \mid Q_i \leq 0 \}, \quad i = 1\dots M
    \label{eq:epsilon_0i}
    \end{equation}
    
    Additionally, to satisfy the inequality $\hat{u}^TD^2\hat{\varphi}(p_c)\hat{u} < 0$, an extra constraint is imposed on $\epsilon$:
    \begin{equation*}
        \epsilon < \frac{-2Q_i\bar{\beta_i}^2}{\frac{1}{2} \bar{\beta_i}\nabla{\bar{\beta_i}} \cdot \nabla{\gamma_d} + \gamma_d \hat{u}^T \Bigg[ \left( 1 - \frac{1}{k} \right) {\nabla\bar{\beta_i}} \nabla{\bar{\beta_i}}^T -  \bar{\beta_i}D^2\bar{\beta_i}\Bigg] \hat{u}}
    \end{equation*}
    
    A sufficient condition for this constraint is:
    \begin{equation*}
        \epsilon < \frac{\min\limits_{\mathcal{B}_i(\epsilon)}\{2|Q_i|\bar{\beta_i}^2\}}{\max\limits_{\mathcal{B}_i(\epsilon)}\left\{\frac{1}{2}\bar{\beta_i} \nabla{\bar{\beta_i}} \cdot \nabla{\gamma_d} + \gamma_d \hat{u}^T \Bigg[ \left( 1 - \frac{1}{k} \right) {\nabla\bar{\beta_i}} \nabla{\bar{\beta_i}}^T - \bar{\beta_i}D^2\bar{\beta_i} \Bigg] \hat{u} \right\}}
    \end{equation*}
    
    Now, let us interpret the right-hand side of the above inequality as a scalar-valued function $\zeta(\epsilon)$. Given that $\epsilon < \epsilon'$, we have $\mathcal{B}_i(\epsilon) \subset \mathcal{B}_i(\epsilon')$, which leads to the inequality $\zeta(\epsilon) \geq \zeta(\epsilon')$. Therefore, it is sufficient to ensure that
    \begin{equation}
        \epsilon < \frac{\min\limits_{\mathcal{B}_i(\epsilon'_{0i})}\{2|Q_i|\bar{\beta_i}^2\}}{\max\limits_{\mathcal{B}_i(\epsilon'_{0i})}\left\{\frac{1}{2}\bar{\beta_i} \nabla{\bar{\beta_i}} \cdot \nabla{\gamma_d} + \gamma_d \hat{u}^T \Bigg[ \left( 1 - \frac{1}{k} \right) {\nabla\bar{\beta_i}} \nabla{\bar{\beta_i}}^T - \bar{\beta_i}D^2\bar{\beta_i} \Bigg] \hat{u} \right\}} \triangleq \epsilon''_{0i}
        \label{eq:eps_0i2}
    \end{equation}
    
    Thus, we define
    \begin{equation}
        \epsilon_0 = \min\{\epsilon'_{0i},\epsilon''_{0i}\} \quad i=1\dots m
        \label{eq:eps_0}
    \end{equation}
\end{proof}

\begin{claim}
    If $k \geq N(\epsilon)$ from Eq.~(\ref{eq:N_eps}), then there exists an $\epsilon_1 > 0$ such that $\hat{\varphi}$ has no critical points on $\mathcal{F}_1(\epsilon)$, the region near the workspace boundary, for all $\epsilon < \epsilon_1$.
\end{claim}

\begin{proof}
    Since the boundary of the workspace is a sphere, the proof presented in~\cite{Rimon1988} is directly applicable here.
\end{proof}

Building on the previous result that all critical points must lie in the interior of $\mathcal{F}$, it remains to show that these points are non-degenerate, ensuring that $\hat{\varphi}$ is a Morse function. This is established in~\cite{rimon1988exact}, where it is shown that $\hat{\varphi}$ is Morse if $0 < \epsilon < \epsilon_2$.

Using this result, we complete the theorem by defining $N(\epsilon) = N(\epsilon_{\min})$ in Eq.~(\ref{eq:N_eps}), where
\begin{equation}
    \epsilon_{\min} \triangleq \frac{1}{2} \min\{\epsilon_0, \epsilon_1, \epsilon_2\}
\end{equation}

%%%%%%%%%%%%%%%%%%%%%%%%%%%%%%%%%%%
\subsection{Intersecting Obstacles}
\label{sec:intersecting_obstacles}
%%%%%%%%%%%%%%%%%%%%%%%%%%%%%%%%%%%

\begin{figure}[t!]
    \centering
    \includegraphics[width=0.8\linewidth]{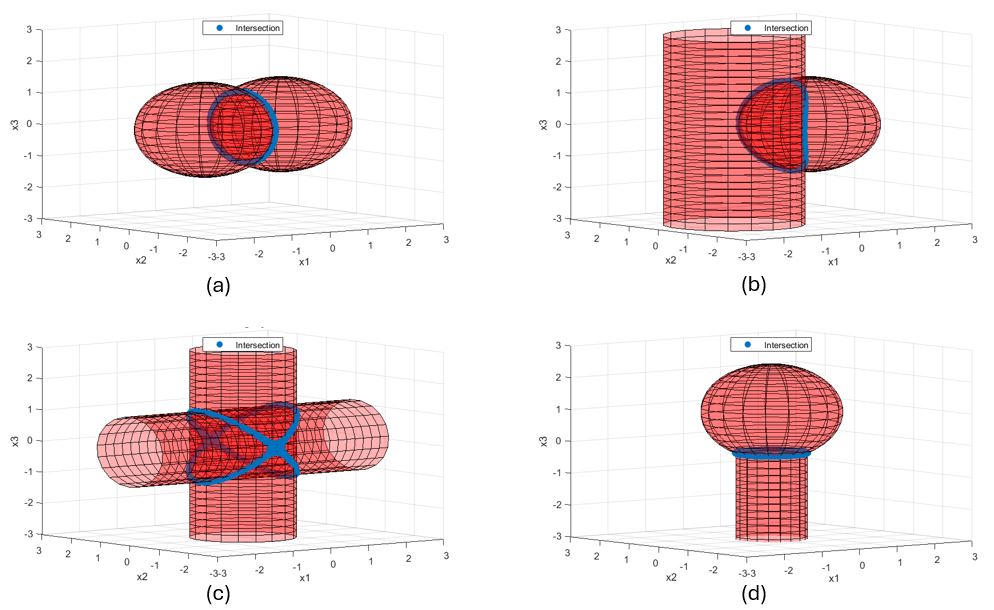}
    \caption{Examples of intersecting obstacles: (a) Intersecting spheres. (b) Intersecting sphere and a cylinder. (c) Intersecting perpendicular cylinders. (d) Intersecting half-cylinder and sphere.}
    \label{fig:intersecting_obst}
\end{figure}

\begin{figure}[!ht]
    \centering
    \includegraphics[width=0.8\linewidth]{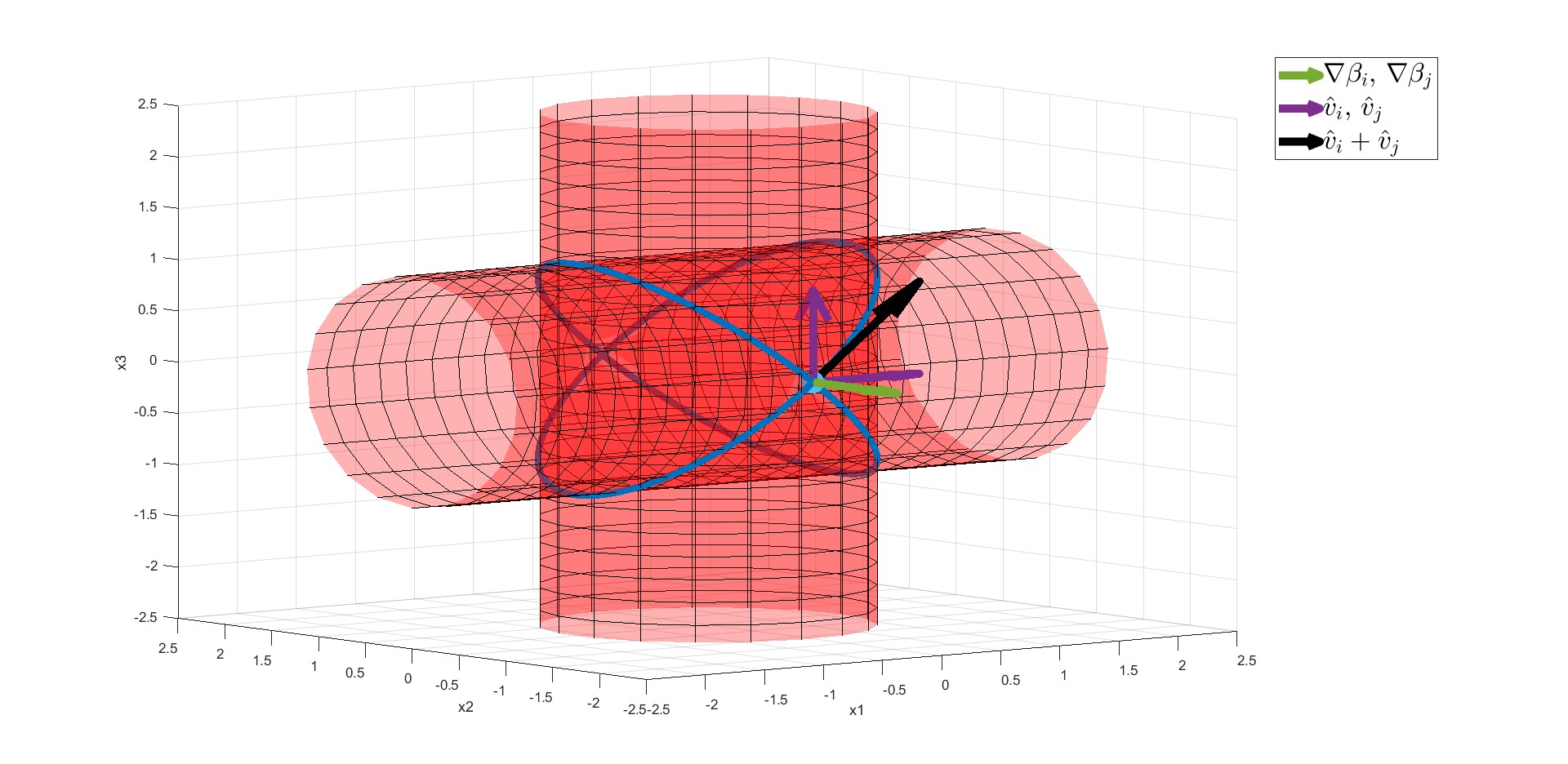}
    \caption{Illustration of the point along the intersection of two cylinders where $\nabla\beta_i\parallel\nabla\beta_j$. The vector $\hat{v}_i+\hat{v}_j$ shown in black is perpendicular to the gradient vector shown in green.}
    \label{fig:intersecting_cylinder}
\end{figure}

In our workspace, we only permit \textit{pairwise unions} of obstacles from the obstacle catalog defined in Section~\ref{sec:base_nav_fun}. Fig.~\ref{fig:intersecting_obst} illustrates different types of allowed obstacle intersections, with the intersection curves highlighted in blue. In Fig.~\ref{fig:intersecting_obst}(a), two intersecting spheres form a circular intersection curve. Fig.~\ref{fig:intersecting_obst}(b) depicts the intersection between a sphere and a cylinder, where the intersection curve takes the shape of an arched ellipse. Fig.~\ref{fig:intersecting_obst}(c) shows two intersecting perpendicular cylinders, producing two planar elliptical intersection curves (Steinmetz curve~\cite{gray1997modern}). For simplicity, in addition to the constraint of perpendicular cylinders, we also demand that both cylinders have equal radii and that their axes intersect. Lastly, Fig.~\ref{fig:intersecting_obst}(d) depicts the intersection of a half-cylinder with a sphere. Note that when the center of the sphere lies on the cylinder's axis the intersection curve is a perpendicular slice of the cylinder. 

While most of the proofs from previous sections remain valid in this scenario, it is necessary to re-evaluate the constraint on $\epsilon_0$ to ensure that no local minima appear near the intersection. Let us first examine the case of intersecting cylinders. Consider the line passing through the intersection of both axes and perpendicular to $\hat{v}_i$ and $\hat{v}_j$. On this line, we have $\nabla \beta_i \parallel \nabla \beta_j$. An illustration of the points $p_s$ on the intersection curve where $\nabla \beta_i(p_s) \parallel \nabla \beta_j(p_s)$ is shown in Fig.~\ref{fig:intersecting_cylinder}. We first show that at $p_s$, the vector sum $\hat{v}_i + \hat{v}_j$ is perpendicular to both $\nabla \beta_i(p_s)$ and $\nabla \beta_j(p_s)$. This result will be useful in the subsequent proof.

Let $p_i = p_j$, meaning that $p_i$ and $p_j$ represent the intersection point of the cylinders' axes. Then,  
\begin{equation*}
    \begin{aligned}    
        \nabla \beta_i(p_s) \cdot (\hat{v}_i + \hat{v}_j) &= \nabla \beta_i(p_s) \cdot \hat{v}_i + \nabla \beta_i(p_s) \cdot \hat{v}_j \\
        &= 2 \left( (p_s - p_i) - \left(\hat{v}_i\cdot (p_s - p_i)\right) \hat{v}_i \right) \cdot \hat{v}_i \\
        &\quad + 2 \left( (p_s - p_i) - \left(\hat{v}_i\cdot (p_s - p_i)\right) \hat{v}_i \right) \cdot \hat{v}_j.
    \end{aligned}
\end{equation*}

\noindent Since $(p_s - p_i) \perp \hat{v}_i$ and $(p_s - p_i) \perp \hat{v}_j$, we obtain  
\begin{equation*}
    \nabla \beta_i(p_s) \cdot (\hat{v}_i + \hat{v}_j) = 0.
\end{equation*}

\begin{figure}[t!]
    \centering
    \includegraphics[width=0.8\linewidth]{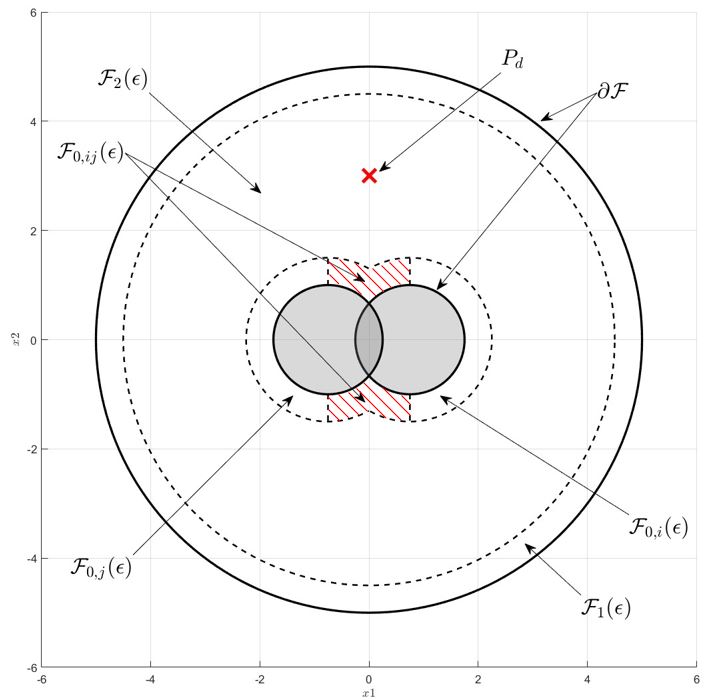}
    \caption{Illustration of $\mathcal{F}_{0,ij}(\epsilon)$, the region near the intersection of two obstacles.}
    \label{fig:intersection_region}
\end{figure}

\noindent Note that since $\nabla \beta_i(p_s) \parallel \nabla \beta_j(p_s)$, if $\nabla \beta_i(p_s) \perp (v_i+v_j)$ then also $\nabla \beta_j(p_s) \perp (v_i+v_j)$. Now, similarly to the regions defined in Section~\ref{sec:unique_min_proof}, we define $\mathcal{F}_{0,ij}(\epsilon)$ as the region near the intersection of the $i$-th and $j$-th obstacles. An illustration of this region is shown in Fig.~\ref{fig:intersection_region}.

\begin{claim}
     For two intersecting obstacles, there exists $\epsilon_0 > 0$ such that $\hat{\varphi}$ has no local minima in $\mathcal{F}_{0,ij}(\epsilon)$, the region near the intersection, for all $\epsilon < \epsilon_0$.
\end{claim}

\begin{figure}
    \centering
    \includegraphics[width=\linewidth]{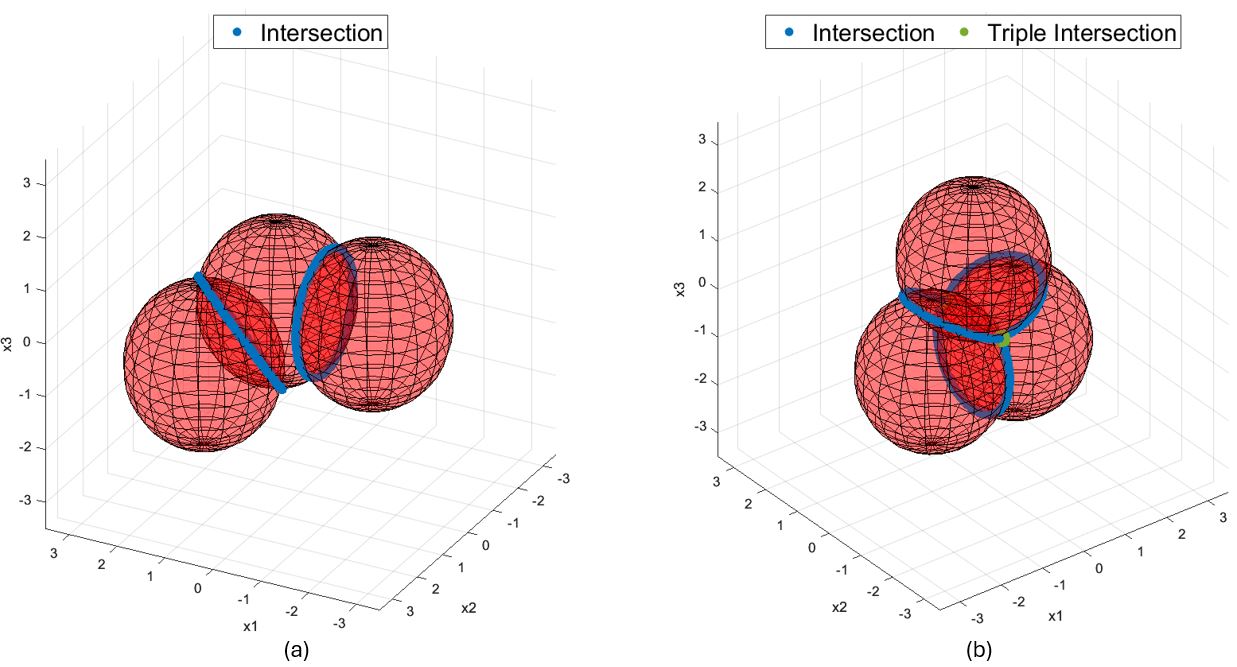}
    \caption{Illustration of three intersecting spherical obstacles. (a) Two separate intersections. (b) A triple intersection. The blue curves represent the pairwise intersections of two spheres, while the green points indicate the triple intersection, which resides in a concave region.}
    \label{fig:three_sphere_int}
\end{figure}

\begin{proof}
    Substituting for $\nabla{\gamma_d} \nabla\gamma_d^T$ in Eq.~(\ref{eq:hessian_at_Cp}) using Eq.~(\ref{eq:grad_gamma_product}), and employing the relation $\beta = \beta_{ij} \bar{\beta}_{ij}$, where $\beta_{ij}=\beta_i\beta_j$, we obtain  

    \begin{align*}
        {D}^2\hat{\varphi}(p_c) = \frac{{\gamma_d}^{k-1}}{\beta^2} &\Bigg( k\beta {D}^2\gamma_d + \left( 1-\frac{1}{k} \right) \frac{\gamma_d}{\beta} \Big[ \beta_{ij}^2 \nabla{\bar{\beta}_{ij}} \nabla\bar{\beta}_{ij}^T \\
        &+ 2\beta_{ij} \bar{\beta}_{ij}(2\nabla{\bar{\beta}_{ij}} \nabla{\beta}_{ij}^T) + \bar{\beta}_{ij}^2 \nabla{\beta_{ij}} \nabla\beta_{ij}^T \Big] \\
        &-\gamma_d \Big[ \beta_{ij} D^2 \bar{\beta}_{ij} + 2(\nabla \bar{\beta}_{ij}^T  \nabla{\beta_{ij}}) + \bar{\beta}_{ij} D^2 {\beta_{ij}} \Big] \Bigg)
    \end{align*}

    \noindent Now, we evaluate the quadratic form associated with $ {D}^2\hat{\varphi}(p_c)$ at  $\hat{u}$ where,$\hat{u}\perp\nabla{\beta_{i}}$ and $\hat{u}\perp\nabla{\beta_{j}}$. Note that for most points it is sufficient to define $\hat{u} = \widehat{\nabla{\beta_{i}\times{\nabla\beta_j}}}$. However, when $\nabla{\beta_{i}} \| \nabla{\beta_{j}}$ which can only happen for two intersecting cylinders, as shown in Fig.~\ref{fig:intersecting_cylinder}, we define $\hat{u} = \widehat{\hat{v}_i+\hat{v}_j}$. Since $D^2\gamma_d=2I$ and $D^2\beta_{ij}=2(\beta_i+\beta_j)I+2\nabla\beta_i{\nabla\beta_j}^T$, the resulting quadratic form is 

    \begin{equation}
    \begin{aligned}
         \frac{\beta^2}{{\gamma_d}^{k-1}} \hat{u}^T \left[D^2\hat{\varphi}(p_c)\right] \hat{u} &= 2k\beta - 2\gamma_d(\bar{\beta_i}+\bar{\beta_j}) \\ 
        &+\hat{u}^T \Bigg[ \left( 1 - \frac{1}{k} \right) \frac{\gamma_d}{\beta} {\beta_{ij}}^2 {\nabla\bar{\beta}_{ij}} \nabla\bar{\beta}_{ij}^T - \gamma_d \beta_{ij} D^2\bar{\beta}_{ij} \Bigg]\hat{u}
    \end{aligned}
    \label{eq:q_form_ss}
    \end{equation}

Considering the inner product of Eq.~(\ref{eq:grad_at_Cp}) with $\nabla{\gamma_d}$ we obtain

\begin{equation}
    \begin{aligned}
        4k\beta = \nabla{\beta} \cdot \nabla{\gamma_d} &= \bar{\beta}_{ij} \left(\beta_i\nabla\beta_j + \beta_j\nabla\beta_i   \right) \cdot \nabla{\gamma_d} + \beta_{ij} \nabla{\bar{\beta}_{ij}} \cdot 
        \nabla{\gamma_d} \\ &=  \left(\bar{\beta_{j}}\nabla\beta_j + \bar{\beta_{i}}\nabla\beta_i \right) \cdot \nabla{\gamma_d} + \beta_{ij} \nabla{\bar{\beta}_{ij}} \cdot 
        \nabla{\gamma_d}
    \end{aligned}    
\end{equation}

 Substituting for $2k\beta$ in Eq.~(\ref{eq:q_form_ss}) results in the following equation
    \begin{equation}
    \begin{aligned}
         \frac{\beta^2}{{\gamma_d}^{k-1}} \hat{u}^T [D^2\hat{\varphi}&(p_c)] \hat{u} = 2\bar{\beta_i} \left( \frac{1}{4}\nabla{\beta_i} \cdot \nabla{\gamma_d} - \gamma_d \right) +  2\bar{\beta_j} \left( \frac{1}{4}\nabla{\beta_j} \cdot \nabla{\gamma_d} - \gamma_d \right) \\ 
        &+ \beta_{ij}\left( \frac{1}{2} \nabla{\bar{\beta}_{ij}} \cdot \nabla{\gamma_d} + \gamma_d \hat{u}^T \Bigg[ \left( 1 - \frac{1}{k} \right) \frac{1}{\bar{\beta}_{ij}} {\nabla\bar{\beta}_{ij}} \nabla\bar{\beta}_{ij}^T - D^2\bar{\beta}_{ij} \Bigg] \hat{u} \right)
    \end{aligned}
    \label{eq:q_form_ss_final}
    \end{equation}

\noindent Noting that the first and second terms in Eq.~(\ref{eq:q_form_ss_final}) correspond to $Q_i$ and $Q_j$ from Eq.~(\ref{eq:ui_cylinder}), we obtain a condition similar to Eq.~(\ref{eq:epsilon_0i}). The constraint on $\epsilon$ is therefore given by

\begin{equation*}
    \epsilon < \epsilon'_{0l} \triangleq \max \{ \epsilon \mid Q_i \leq 0 \wedge Q_j \leq 0\}, \quad l=1\dots m_{int}
\end{equation*}

\noindent Where $m_{int}$ is the number of intersecting obstacle pairs. Note that a single obstacle can intersect with multiple other obstacle as long as there is no triple intersection as illustrated in Fig.~\ref{fig:three_sphere_int}(b). Therefore, using Eq.~(\ref{eq:q_form_ss_final}) we can modify Eq.~(\ref{eq:eps_0i2}) to be 
\begin{equation}
        \epsilon < \frac{\min\limits_{\mathcal{B}_{ij}(\epsilon'_{0l})}\{2\bar{\beta}_{ij}^2\left(|Q_i|\beta_j + |Q_j|\beta_i\right)\}}{\max\limits_{\mathcal{B}_{ij}(\epsilon'_{0l})}\left\{\frac{1}{2}\bar{\beta}_{ij} \nabla{\bar{\beta}_{ij}} \cdot \nabla{\gamma_d} + \gamma_d \hat{u}^T \Bigg[ \left( 1 - \frac{1}{k} \right) {\nabla\bar{\beta}_{ij}} \nabla\bar{\beta}_{ij}^T - \bar{\beta}_{ij}D^2\bar{\beta}_{ij} \Bigg] \hat{u} \right\}} \triangleq \epsilon''_{0l}
    \end{equation}

Thus we define
\begin{equation}
        \epsilon_{0,int} = \min\{\epsilon'_{0l},\epsilon''_{0l}\} \quad l=1\dots m_{int}
        \label{eq:eps_0_int}
    \end{equation}

\end{proof}

Let us now modify $\epsilon_0$ in Section~\ref{sec:interior_F} to be the minimum between $\epsilon_{0,int}$ from Eq.~(\ref{eq:eps_0}) and $\epsilon_0$ from Eq.~(\ref{eq:eps_0_int}) thus expanding the proof that their are no local minima to include pairwise unions of obstacles. 

\textbf{Caveat with triple intersections:} When three obstacles intersect, concave regions may form near the \textit{triple intersection}, as illustrated in Fig.~\ref{fig:three_sphere_int}(b). The test direction $\hat{u}$, which facilitates escape from such cavities must be perpendicular to the repulsion gradient of each obstacle. Consequently, at a triple intersection point no valid escape direction exists and local minima may form in these regions. A formal proof is shown in~\cite{filippidis2013navigation}. Therefore, triple intersections are \textit{not} permitted in our workspace. \hfill$\circ$

While our 3-D model workspace does not admit triple intersections. they can be handled using coordinate transformation or workspace decomposition in future work.

%%%%%%%%%%%%%%%%%%%%%%%%%%%%%%%
%%%%%%%%%%%%%%%%%%%%%%%%%%%%%%%
\section{Simulated Environments}
\label{sec:simulations}
%%%%%%%%%%%%%%%%%%%%%%%%%%%%%%%
%%%%%%%%%%%%%%%%%%%%%%%%%%%%%%%
This section evaluates the effectiveness of our navigation functions through simulations. Section~\ref{sec:general_sim} simulates a general workspace with ten randomly placed internal obstacles. Section~\ref{sec:truss_sim} simulates navigation in a workspace containing a truss obstacle, created by combining the base obstacles. Recall from Section~\ref{sec:composite_vav_func} that our navigation functions are:

\begin{equation*}
    \varphi(\vec{x}) = \sigma_d \circ \sigma \circ \hat{\varphi}(\vec{x}) = \frac{\gamma_d}{({\gamma_d}^k + \beta)^{1/k}} 
\end{equation*}
\begin{equation*}
    \psi(\vec{x}) = \sigma \circ \sigma_d \circ \hat{\varphi}(\vec{x})  = \frac{\gamma_d}{\gamma_d + \beta^{1/k}}
\end{equation*}

%%%%%%%%%%%%%%%%%%%%%%%%%%%%%%%%
\subsection{General Workspace}
\label{sec:general_sim}
%%%%%%%%%%%%%%%%%%%%%%%%%%%%%%%%

\begin{figure}[t!]
    \centering
    \includegraphics[width=0.85\linewidth]{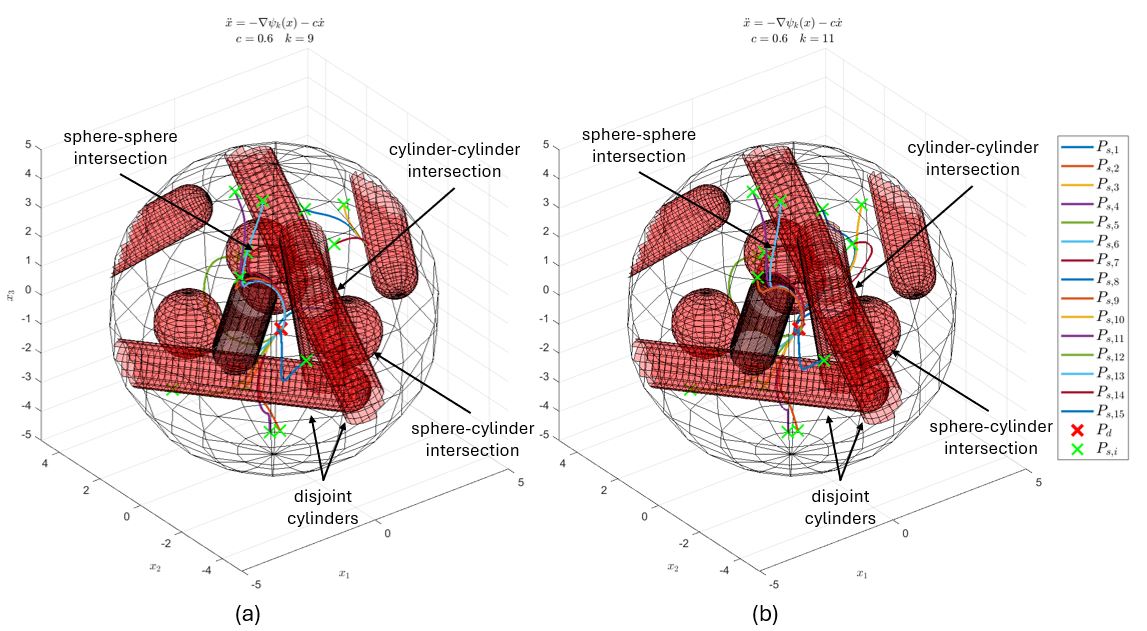}
    \caption{Example simulation of the $\psi$ navigation function with a damping coefficient of $c=0.6$. (a) For tuning parameter $k=9$, three initial points converge to local minima. (b) For tuning parameter $k=11$, all evaluated initial points successfully reach the destination.}
    \label{fig:nav_map_rand}
\end{figure}

\begin{figure}[t!]
    \centering
    \includegraphics[width=0.85\linewidth]{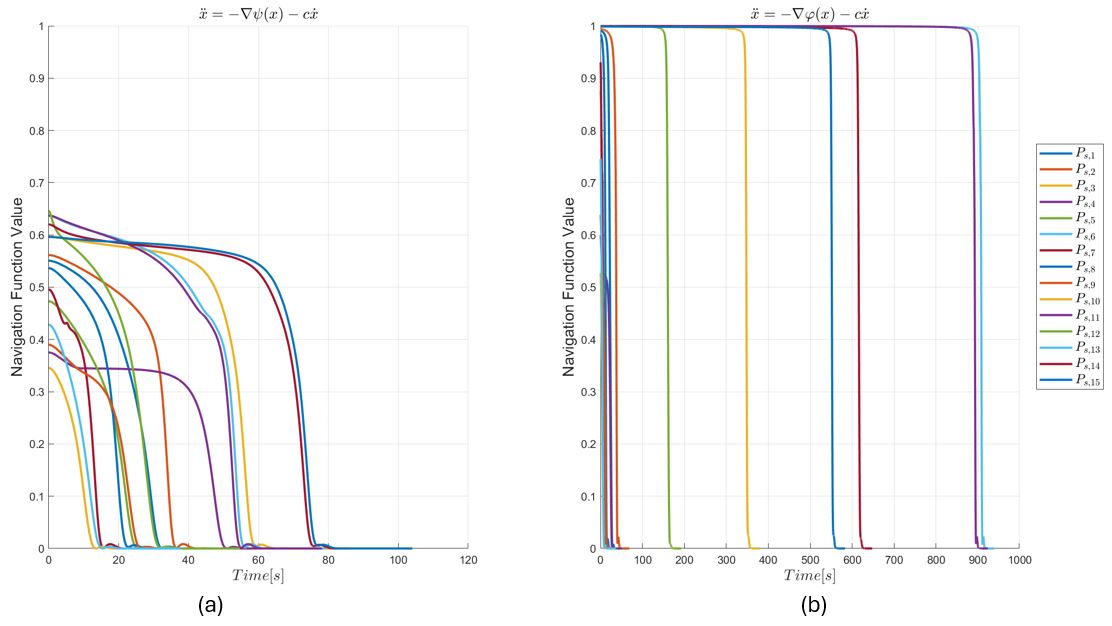}
    \caption{Navigation function value over time for tuning parameter $k=11$. (a) values for $\psi$. (b) values for $\varphi$.}
    \label{fig:nav_val_rand}
\end{figure}

To evaluate the effectiveness of the navigation functions, we conducted simulations in ten distinct workspaces. Each workspace, was defined with a radius of $r_0=5~m$ and contains ten randomly placed internal obstacles from the catalog defined in Section~\ref{sec:base_nav_fun}. Note that obstacles may be either disjoint or pairwise intersecting, as specified in Section~\ref{sec:intersecting_obstacles}.  
For each workspace we performed ten simulations, each with a different destination point, resulting in a total of 100 simulations. In each workspace, 15 initial points were randomly selected. Each destination point was assessed by navigating from these 15 initial points to the target.

The robot's control law was defined as  
\begin{equation}
    \ddot{x} = -\nabla\lambda(x) - c\dot{x}
    \label{eq:control_law}
\end{equation}
where $\lambda$ is the navigation function ($\varphi$ or $\psi$), $x$ represents the robot's position, and $c$ is the damping coefficient.

\begin{figure}[!t]
    \centering
    \includegraphics[width=0.5\linewidth]{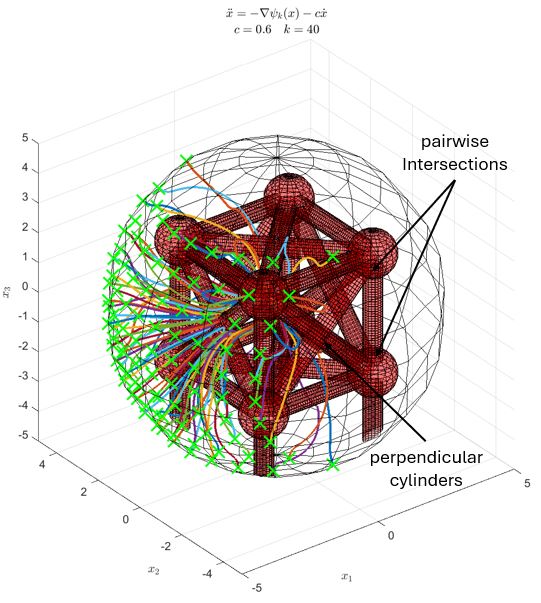}
    \caption{Trajectories for simulation of a truss obstacle for 100 initial positions and common target at $P_d=(0,0,0)$.}
    \label{fig:nav_map_truss}
\end{figure}

An example simulation result for the $\psi$ function with $k=9$ and $c=0.6$ is presented in Fig.~\ref{fig:nav_map_rand}(a). The results for the same workspace with $k=11$ are shown in Fig.~\ref{fig:nav_map_rand}(b). It is evident that for $k=9$, the destination was not reached for three initial positions, as the robot converged to a local minimum. In contrast, for $k=11$, all evaluated initial positions successfully reached the destination. The evolution of the navigation function values over time for the simulation in Fig.~\ref{fig:nav_map_rand}(b) is depicted in Fig.~\ref{fig:nav_val_rand}.

It is evident that for the $\psi$ function, the destination is reached relatively quickly (within 80 seconds for all initial positions), as shown in Fig.~\ref{fig:nav_val_rand}(a). In contrast, for the $\varphi$ function, the navigation time for some initial positions can exceed 900 seconds, as depicted in Fig.~\ref{fig:nav_val_rand}(b). This behavior can be explained by the structure of the functions. Assuming $\gamma_d>1$, we can consider the following limits:
\begin{gather*}
     \lim_{k \to \infty} \varphi = \frac{\gamma_d}{({\gamma_d}^k+\beta)^{1/k}} \xrightarrow[k \to \infty]{} 1 \\
     \lim_{k \to \infty} \psi = \frac{\gamma_d}{{\gamma_d}+\beta^{1/k}} \xrightarrow[k \to \infty]{} \frac{\gamma_d}{{\gamma_d}+1}
\end{gather*}

It is evident that increasing $k$ flattens $\varphi$, whereas $\psi$ retains its convex shape as $k$ increases. Consequently, when the initial state lies in a near-flat region where $\varphi \approx 1$, the robot drifts slowly until it exits this plateau. In environments with a large number of obstacles, larger values of $k$ are required to prevent local minima, further limiting the effectiveness of $\varphi$. Although normalizing the dimensions of the workspace to $r_0 = 0.5$ could mitigate this issue, we will focus exclusively on the function $\psi$ in future evaluations.

%%%%%%%%%%%%%%%%%%%%%%%%%%%%%%
\subsection{Composite Truss Obstacle}
\label{sec:truss_sim}
%%%%%%%%%%%%%%%%%%%%%%%%%%%%%%

\begin{figure}[t!]
    \centering
    \includegraphics[width=0.85\linewidth]{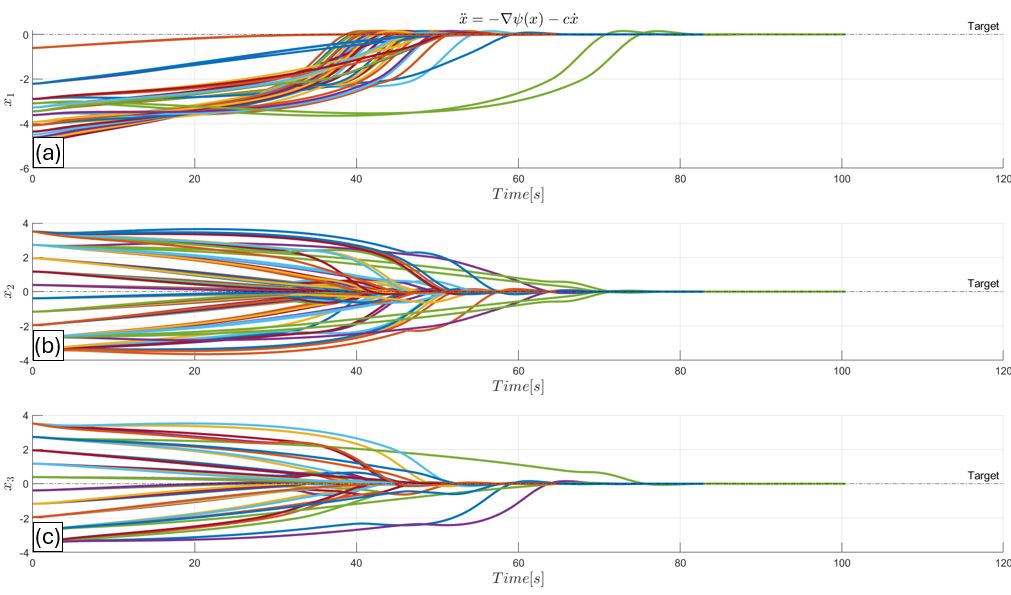}
    \caption{Position of the robot along the trajectory for the truss obstacle simulation. (a) X-coordinate, (b) Y-coordinate and (c) Z-coordinate.}
    \label{fig:nav_pos_truss}
\end{figure}

\begin{figure}[t!]
    \centering
    \includegraphics[width=0.85\linewidth]{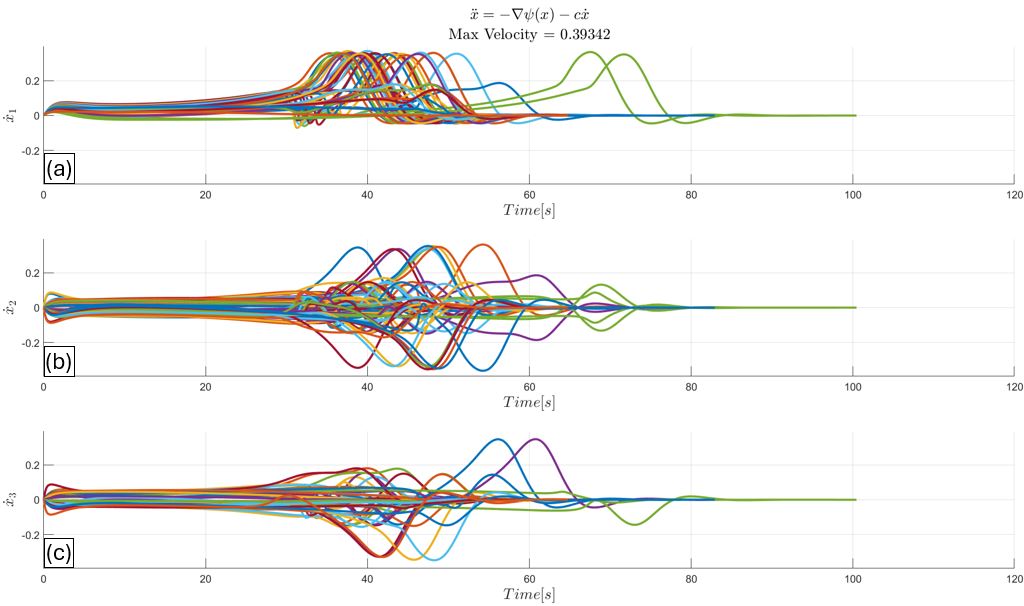}
    \caption{Velocity of the robot along the trajectory for the truss obstacle simulation. (a) X-coordinate, (b) Y-coordinate and (c) Z-coordinate.}
    \label{fig:nav_vel_truss}
\end{figure}

\begin{figure}[ht!]
    \centering
    \includegraphics[width=0.85\linewidth]{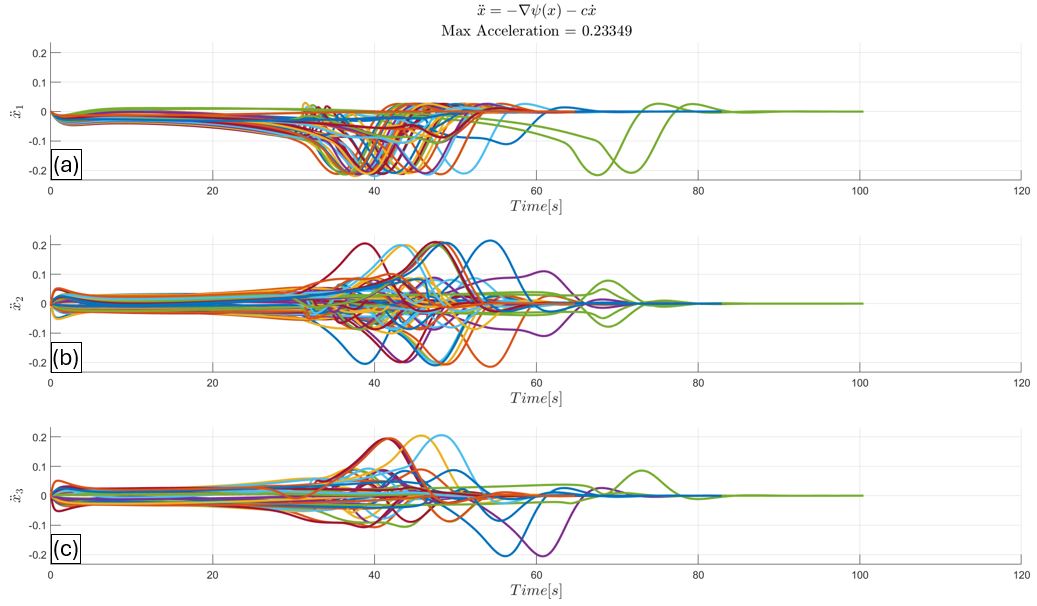}
    \caption{Acceleration of the robot along the trajectory for the truss obstacle simulation. (a) X-coordinate, (b) Y-coordinate and (c) Z-coordinate.}
    \label{fig:nav_acc_truss}
\end{figure}
To construct the truss obstacle depicted in Fig.~\ref{fig:nav_map_truss}, we used 36 base obstacles of spheres, half-cylinders, and finite cylinders. The tuning parameter was set to $k=40$ with a damping coefficient of $c=0.6$. The simulation was conducted for 100 initial positions near the workspace boundary with a common target point at $P_d=(0,0,0)$. The resulting trajectories of the dynamic system (\ref{eq:control_law}) are shown in Fig.~\ref{fig:nav_map_truss}. The position, velocity, and acceleration along these trajectories are depicted in Figs.~\ref{fig:nav_pos_truss} through~\ref{fig:nav_acc_truss}. Notably, the velocity remained below 0.4~$m/s$, while the acceleration did not exceed 0.24~$m/s^2$ at all times along the trajectories.  

\begin{figure}[!t]
    \centering
    \includegraphics[width=\linewidth]{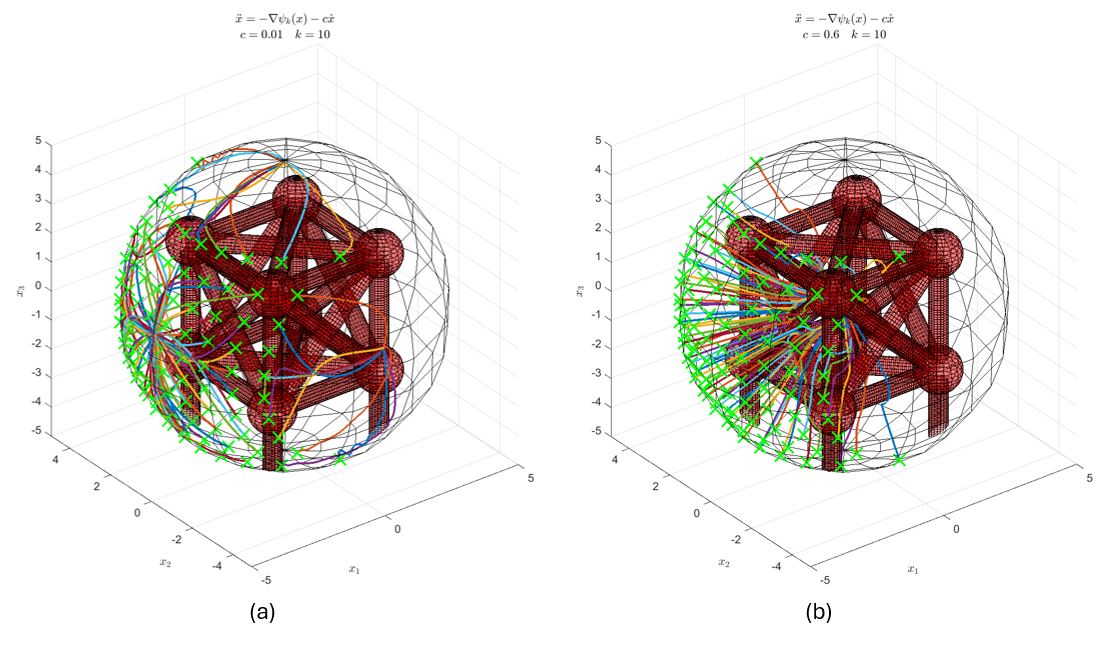}
    \caption{Simulation of the truss obstacle with $k=10$ for 100 initial positions and a common target at $P_d = (0,0,0)$. (a) Without merging the base obstacles, all trajectories converge to a parasitic local minimum. (b) After merging all base obstacles into a single composite obstacle using the \textit{p}-Rvachev function, all initial positions successfully reach the target.}
    \label{fig:truss_rvachev}
\end{figure}

\textbf{Composite obstacles:} Although this obstacle construction method enables straightforward coordinate transformations between the physical world and the 3-D model workspace, it often results in environments containing a large number of obstacles. Consequently, a high tuning parameter $k$ is required to mitigate the risk of local minima. However, the number of internal obstacles can be artificially reduced by merging intersecting obstacles using Rvachev functions~\cite{shapiro1999implicit}, which can help relax the constraint on $k$, as illustrated in Fig.~\ref{fig:truss_rvachev}. Fig.~\ref{fig:truss_rvachev}(a) shows that, for the Truss obstacle with $k=10$ and no merging of the base obstacle, all initial positions converge to parasitic local minima. In contrast, Fig.~\ref{fig:truss_rvachev}(b) demonstrates that after merging all 36 base obstacles into a single composite obstacle, the target is successfully reached from all initial positions. However, it is important to note that merging obstacles may increase the trajectory computation time of the robot's onboard computer.

\begin{figure}[!t]
    \centering
    \includegraphics[width=\linewidth]{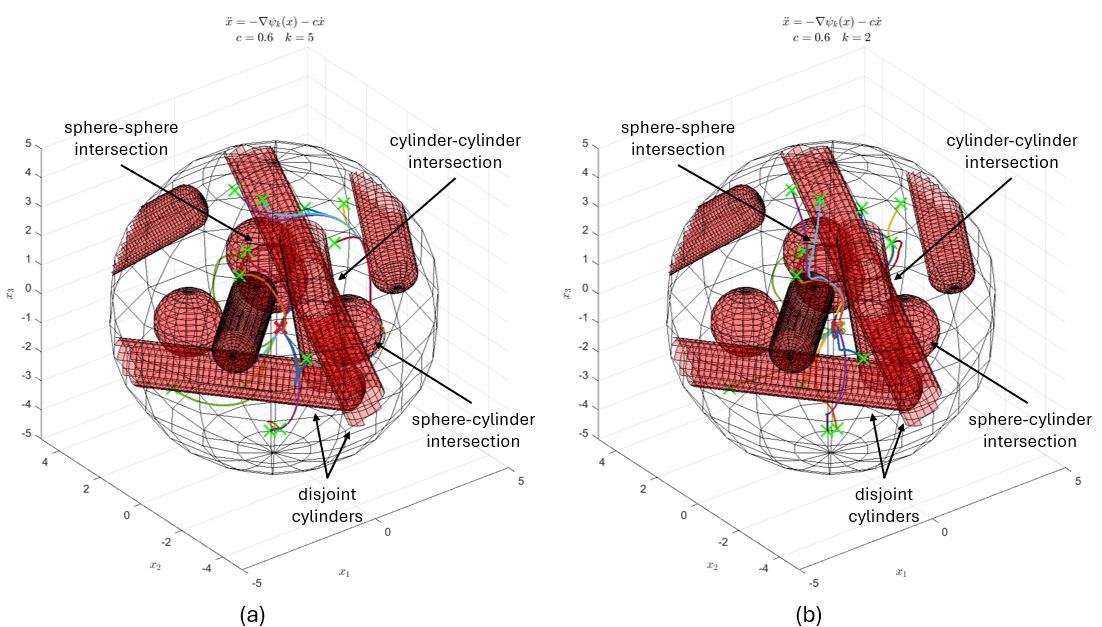}
    \caption{Simulation of a workspace containing ten distinct obstacles. (a) The number of obstacles is reduced to seven by merging only intersecting obstacles; tuning parameter $k=5$ is used. (b) All obstacles are merged into a single obstacle; tuning parameter $k=2$ is used.}
    \label{fig:rand_rvachev}
\end{figure}

To assess the impact of  base obstacle merging in a workspace composed of disjoint obstacles, we re-simulated the scenario from Fig.~\ref{fig:nav_map_rand} using the merging method. Recall that in Fig.~\ref{fig:nav_map_rand}, all ten obstacles were disjoint. The target in Fig.~\ref{fig:nav_map_rand} was reached for all evaluated initial positions only when $k > 9$. After merging only intersecting obstacles, reducing the total obstacle count from ten to seven, this requirement was relaxed to $k > 4$ as shown in Fig.~\ref{fig:rand_rvachev}(a). Furthermore, when applying the \textit{p}-Rvachev function to construct a disjoint union of all internal obstacles in Fig.~\ref{fig:nav_map_rand}, the target was successfully reached even for $k = 2$, as illustrated in Fig.~\ref{fig:rand_rvachev}(b). This improvement may stem from the relaxation of the lower bound on $k$ described in Eq.~(\ref{eq:N_eps}). By reducing the number of obstacles, the term $\sum_{i=0}^{M} \max_{\mathcal{W}}\{\|\nabla\beta_i\|\}$ decreases accordingly, which in turn lowers the theoretical lower bound required for $k$.

%%%%%%%%%%%%%%%%%%%%%%%%%%%%%%%
%%%%%%%%%%%%%%%%%%%%%%%%%%%%%%%
\section{Topological Representation of Base Obstacles}
\label{sec:topology}
%%%%%%%%%%%%%%%%%%%%%%%%%%%%%%%
%%%%%%%%%%%%%%%%%%%%%%%%%%%%%%%

\begin{figure}[!t]
    \centering
    \includegraphics[width=0.9\linewidth]{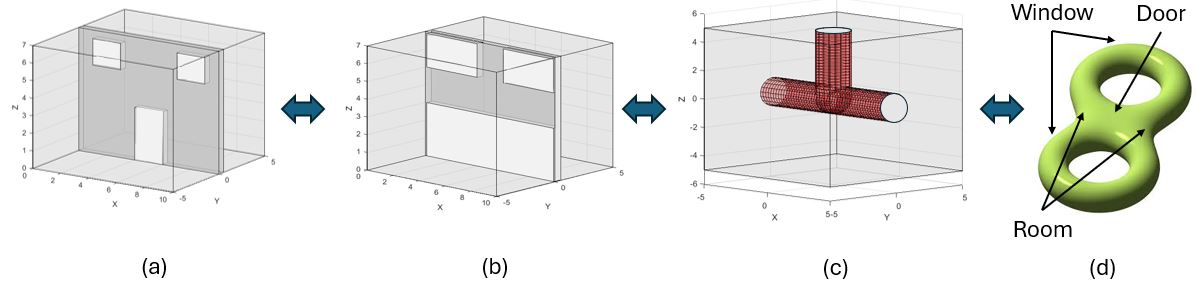}
    \caption{Illustration of a homeomorphism between a physical environment and a 3-D model workspace. (a) A physical environment consisting of two rooms separated by a wall with a door and two windows. (b) A homeomorphic version of the environment, where the openings in the wall have been expanded. (c) A 3-D model workspace representing the homeomorphic environment. (d) Illustration showing that the environment is homeomorphic to the interior of a two-hole torus.}
    \label{fig:homeomorphism}
\end{figure}

In this section, we show via a concrete example that almost any 3-D physical workspace has a homeomorphic counterpart in the 3-D model space, enabling the construction of a coordinate transformation between the two. As illustrated in Fig.~\ref{fig:homeomorphism}(a), we start with a physical environment where two rooms are separated by a wall containing a door and two windows. In Fig.~\ref{fig:homeomorphism}(b) the door and windows are expanded without altering the room's topological structure. As depicted in Fig.~\ref{fig:homeomorphism}(c), the separating wall can now be topologically deformed into an internal obstacle composed of a horizontal full cylinder and a vertical half-cylinder, maintaining the three openings between the two rooms. Notably, this model environment is homeomorphic to the \textit{interior} of a \textit{two-hole torus}, as illustrated in Fig.~\ref{fig:homeomorphism}(d). 

\begin{figure}
    \centering
    \includegraphics[width=0.5\linewidth]{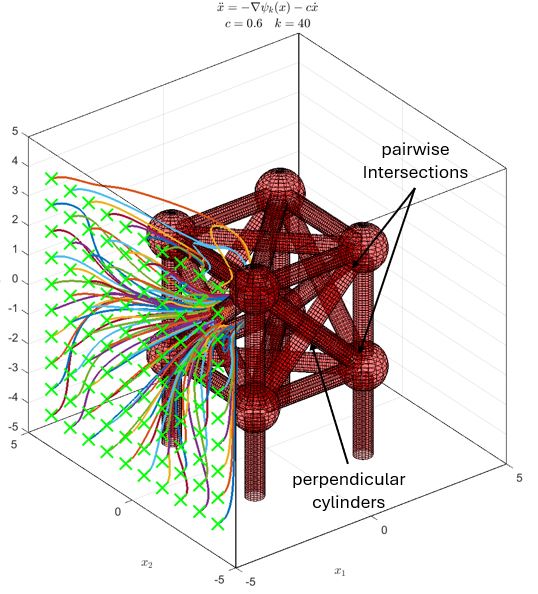}
    \caption{Point-mass robot trajectories for a truss obstacle in a cubic room for 100 initial positions and common target at $P_d=(0,0,0)$ and tuning parameter $k=40$ (due to 36 base obstacles).}
    \label{fig:SQ_truss}
\end{figure}

It is important to note that while we previously used a sphere as the outer boundary of the workspace, the only requirement is that the boundary remains a convex enclosure. Therefore, to simplify the transformation, the outer boundary does not need to be modified in this case. This is further demonstrated in Fig.~\ref{fig:SQ_truss}, where the simulation of the truss obstacle from Section~\ref{sec:truss_sim}, originally conducted in a spherical room, is repeated for a cubic room.

%%%%%%%%%%%%%%%%%%%%%%%%%%%%%%%%%%%%
%%%%%%%%%%%%%%%%%%%%%%%%%%%%%%%%%%%%
\section{Spherical Robot Navigation}
\label{sec:spherical_robot}
%%%%%%%%%%%%%%%%%%%%%%%%%%%%%%%%%%%%
%%%%%%%%%%%%%%%%%%%%%%%%%%%%%%%%%%%%
This section evaluates the navigation of a spherical robot in the 3-D model space. Section~\ref{sec:sphere_point_transform} presents the transformation from the spherical robot to a point robot. Section~\ref{sec:sphere_point_sim} simulates the navigation of a spherical robot.

%%%%%%%%%%%%%%%%%%%%%%%%%%%%%%%%%%%%%%%%%%
\subsection{Transformation to Point Robot}
\label{sec:sphere_point_transform}
%%%%%%%%%%%%%%%%%%%%%%%%%%%%%%%%%%%%%%%%%%
\begin{figure}[!t]
    \centering
    \includegraphics[width=0.9\linewidth]{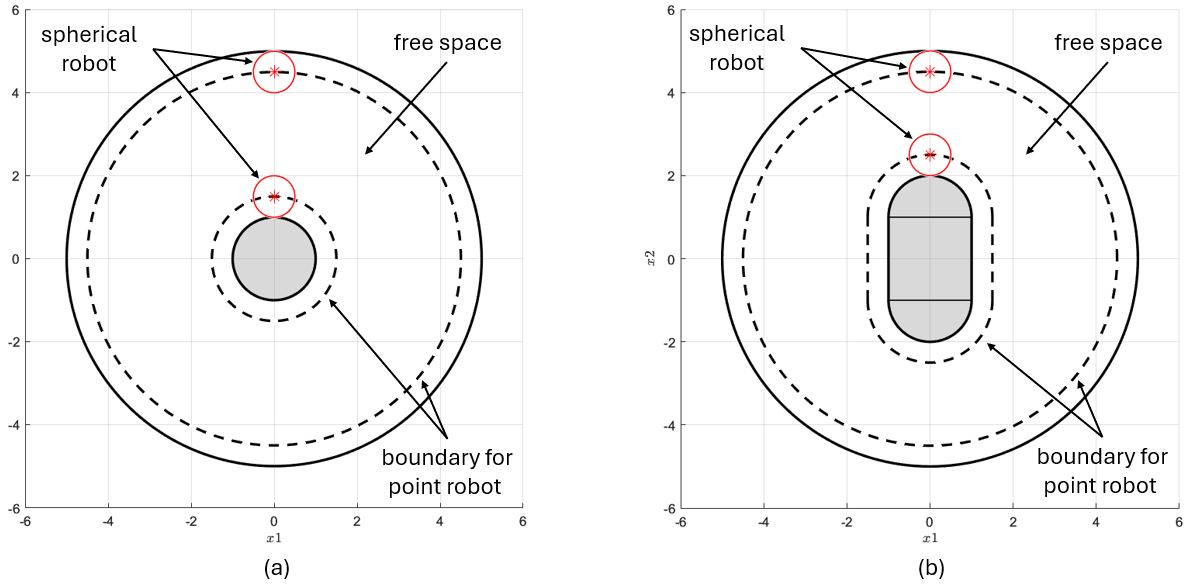}
    \caption{2-D projection of the transformation from spherical robot to point robot for disjoint obstacles: (a) Spherical obstacle. (b) Finite cylindrical obstacle.}
    \label{fig:disjoint_locus}
\end{figure}

\begin{figure}[!t]
    \centering
    \includegraphics[width=0.9\linewidth]{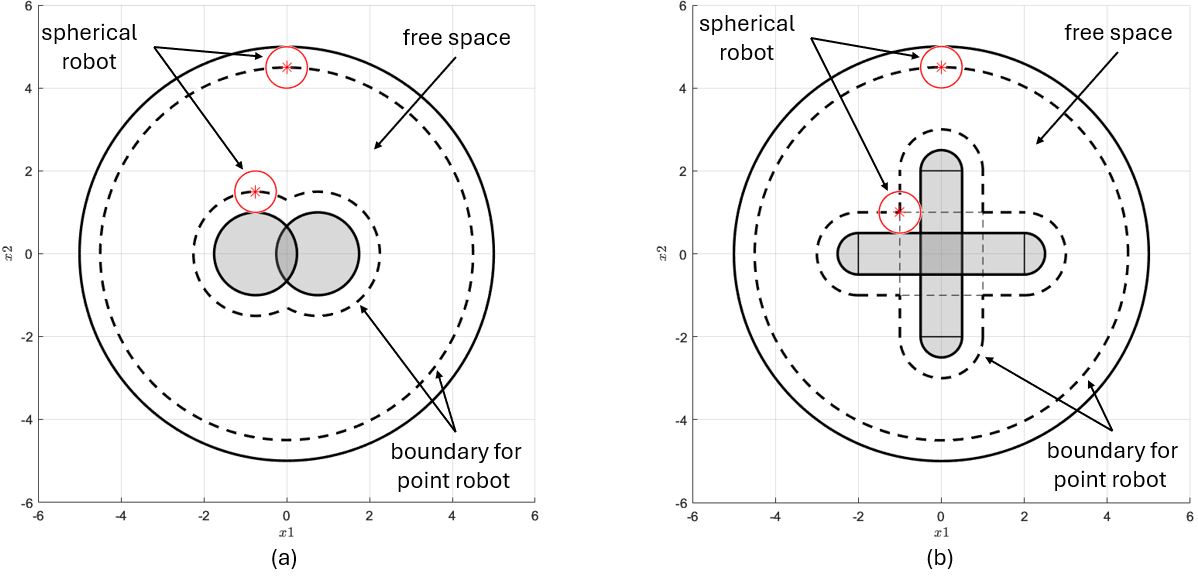}
    \caption{2-D projection of the transformation from spherical robot to point robot for pairwise intersecting obstacles. (a) Intersecting spherical obstacles or sphere and cylinder intersection. (b) Intersection of cylindrical obstacles.}
    \label{fig:intesecting_locus}
\end{figure}

\begin{figure}[!t]
    \centering
    \includegraphics[width=0.9\linewidth]{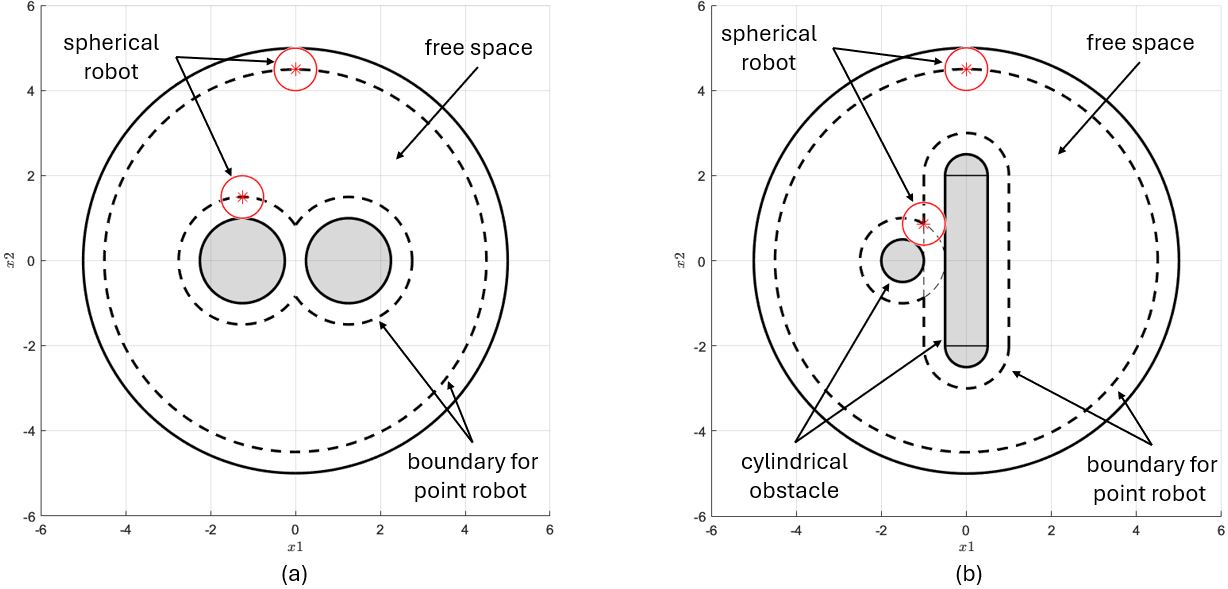}
    \caption{2-D projection of the transformation from spherical robot to point robot for (a) Closely spaced spherical obstacles or sphere and cylinder obstacles, and (b) Closely spaced cylindrical obstacles.}
    \label{fig:near_locus}
\end{figure}

\begin{figure}[!t]
    \centering
    \includegraphics[width=0.9\linewidth]{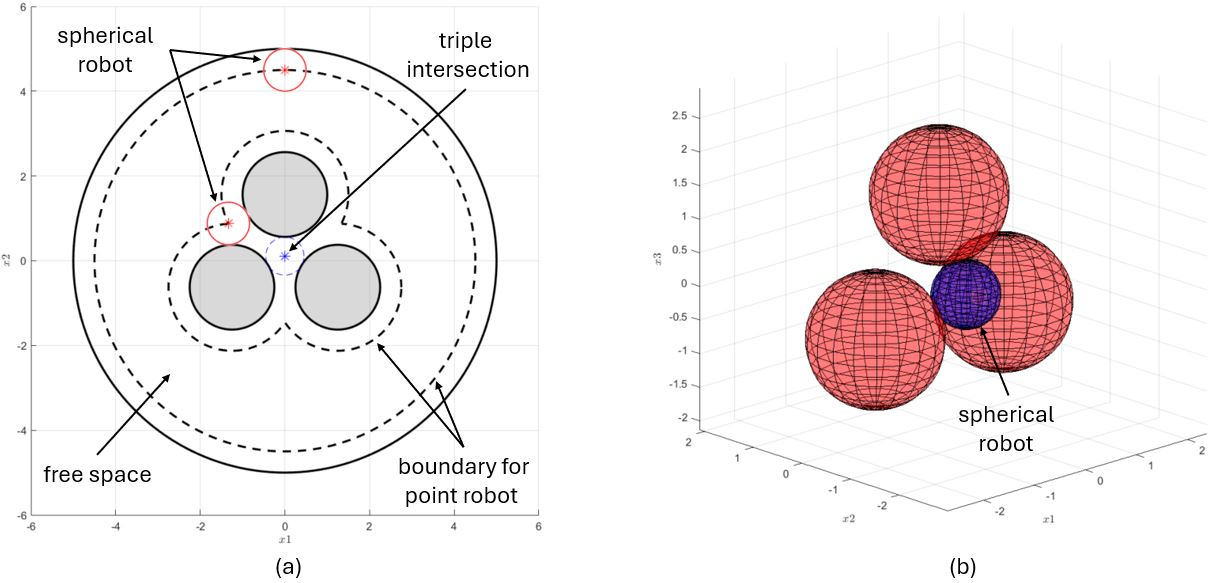}
    \caption{Illustration of the transformation from spherical robot to point robot for obstacle clusters. (a) 2-D projection of the transformation for three closely spaced spherical obstacles. (b) Triple intersection point for a spherical robot shown in the original 3-D model space.}
    \label{fig:cluster_locus}
\end{figure}

Up to this point, we have demonstrated the safe motion planning for a \textit{point-mass robot} within a 3-D model workspace. In this section, we show that a given 3-D model space with a spherical robot can be transformed into an equivalent 3-D model space for a point-mass robot in an obstacle environment. 

The spherical-to-point robot transformation shifts the workspace boundary to represent the locus of the spherical robot \textit{centers} for configurations at which the spherical robot is in contact with the original boundary. This approach effectively shrinks the spherical robot to a point located at its center while expanding the surrounding obstacles by the radius of the spherical robot. As shown in Fig.~\ref{fig:disjoint_locus}(a)-(b) for a spherical and a finite cylindrical obstacle, a robot of radius $R$ results in the radius of each internal obstacle increasing to $r_i' = r_i + R$. Similarly, the boundary of the workspace contracts uniformly, yielding $r_0' = r_0 - R$.

Similarly, Fig.~\ref{fig:intesecting_locus} illustrates the spherical-to-point robot transformation in the case of intersecting obstacles. Fig.~\ref{fig:intesecting_locus}(a) shows the transformation for two intersecting spherical obstacles (or between a sphere and a cylinder), while Fig.~\ref{fig:intesecting_locus}(b) shows the case of intersecting finite cylinders. It can be observed that, for allowed obstacle intersections as described in Section~\ref{sec:intersecting_obstacles}, the transformation preserves the validity of the intersection, thus preserving the original 3-D model space construction. However, in the case of a ball joint, slight modifications are required, as will be discussed later in this section.

\textbf{Closely spaced obstacle pair:} Consider the case where the distance between two obstacles is less than $2R$, as illustrated in Fig.~\ref{fig:near_locus}. In this case, the spherical robot may simultaneously contact both obstacles in certain configurations. Fig.~\ref{fig:near_locus}(a) shows the transformation for closely spaced spherical obstacles (or a sphere and a cylinder), while Fig.~\ref{fig:near_locus}(b) illustrates the case for closely spaced cylinders that do not meet at a ball joint. While close proximity between two spherical obstacles or between a sphere and a cylinder results in an allowed intersection, closely spaced cylinders \textit{do not} yield an allowed 3-D model space. For instance, Fig.~\ref{fig:near_locus}(b) shows two 3-D expanded cylinders that form a pairwise intersection which is \textit{not} allowed by the 3-D model space. Consequently, in a \textit{spherical-robot 3-D model space}, the distance between any two cylinders that do not meet at a ball joint, must be at least $2R$. However, this constraint can be circumvented by using coordinate transformations or by introducing an additional ball joint obstacle that encloses the intersection curves of the cylinders in the \textit{point-robot workspace} similarly to the truss obstacle in Fig.~\ref{fig:nav_map_truss}.

\textbf{Obstacle clusters:} Consider the case of three or more obstacles where the distance between any two obstacles in the cluster is less than $2R$, as shown in Fig.~\ref{fig:cluster_locus}(a). In such cases, there may be a configuration where the spherical robot is simultaneously touching three obstacles, as demonstrated in Fig.~\ref{fig:cluster_locus}(b), resulting in a triple intersection point in the point-robot workspace. Thus, except at the ball joints which form admissible obstacle clusters, the spherical-robot 3-D model space does not permit these clusters.

\begin{figure}[!t]
    \centering
    \includegraphics[width=0.9\linewidth]{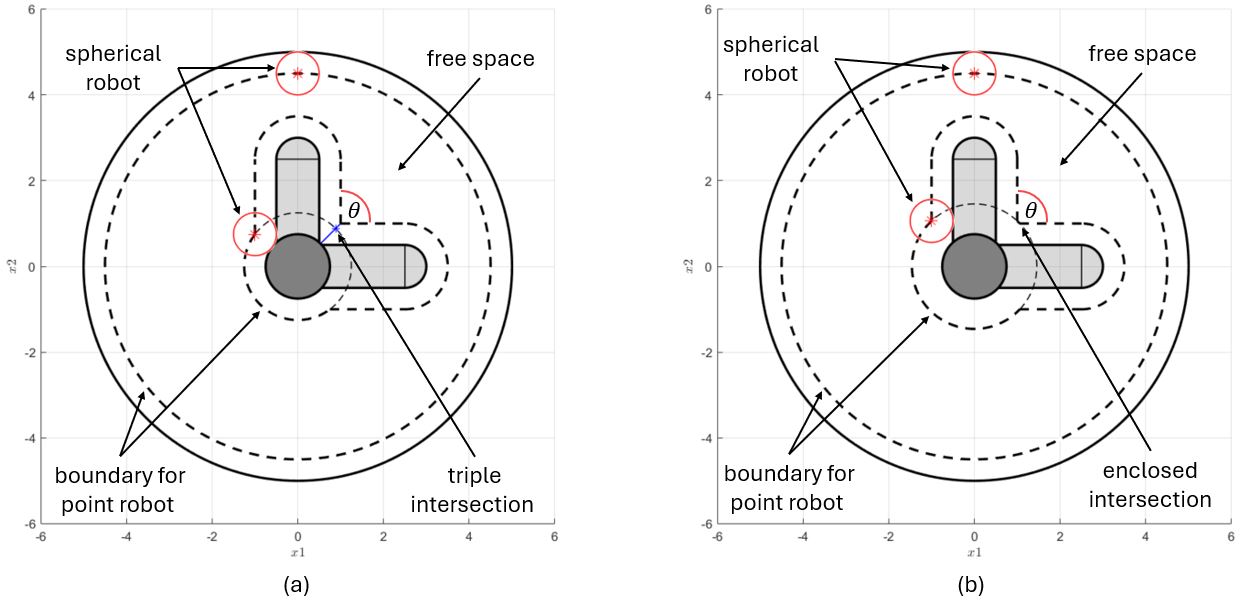}
    \caption{2-D projection of the transformation from spherical robot to point robot at a ball joint. (a) All obstacles are expanded by $R$ resulting in the intersection of the cylindrical obstacles escaping the expanded ball joint. (b) A proper expansion of the ball joint where the intersection of the cylindrical obstacles remains inside the sphere.}
    \label{fig:ball_joint_locus}
\end{figure}

\textbf{Ball joint:} We define a \textit{ball joint} as the intersection of two or more finite cylindrical obstacles enclosed within a sphere, as illustrated by the corners of the truss obstacle in Fig.~\ref{fig:nav_map_truss}. Since the intersection curve of the cylinders may lie arbitrarily close to the surface of the ball joint, the following transformation ensures that the entire intersection remains fully contained within the expanded ball joint obstacle when transforming to the point robot workspace:
\begin{gather}
    r_{\text{cylinder}}' = r_{\text{cylinder}} + R \\
    r_{\text{ball-joint}}' = r_{\text{ball-joint}} + \frac{R}{\sin(\theta/2)}
    \label{eq:fully_enclosed}
\end{gather}

\noindent where $\theta$ denotes the angle between the axes of two intersecting cylinders. For joints involving more than two cylinders, $\theta$ is defined as the minimum angle between any pair of cylinder axes:
\begin{equation*}
    \theta = \min_{i \neq j} \left\{ \arcsin \left( \| \hat{v}_i \times \hat{v}_j \| \right) \right\}.
\end{equation*}

\noindent Smaller values of $\theta$ require a larger expansion of the ball joint to fully enclose the worst-case intersection curve of the cylinders. The additional expansion beyond the physical offset $R$ is given by:
\begin{equation*}
    \Delta = \frac{R}{\sin(\theta/2)} - R = \left( \frac{1}{\sin(\theta/2)} - 1 \right) \cdot R.
\end{equation*}

\noindent For perpendicular cylinders, this additional expansion is approximately $40\%$ ($R/\sin(45^{\circ})-R\cong0.4R$). However, when $R \ll r_{\text{ball-joint}}$, this increase is negligible. The main limitation of the fully enclosed expansion method is that it restricts the placement of a closely spaced obstacle pair within a distance of $\left(\frac{1}{\sin(\theta/2)} + 1\right)R$ from the ball joint, compared to the standard $2R$ clearance for regularly expanded obstacles, as this proximity could lead to unintended triple intersections.

An incorrect transformation where the intersection of two expanded cylinders escapes the ball joint resulting in a triple intersection point is illustrated in Fig.~\ref{fig:ball_joint_locus}(a), while the correct transformation is depicted in Fig.~\ref{fig:ball_joint_locus}(b). 

\begin{figure}[!t]
    \centering
    \includegraphics[width=0.9\linewidth]{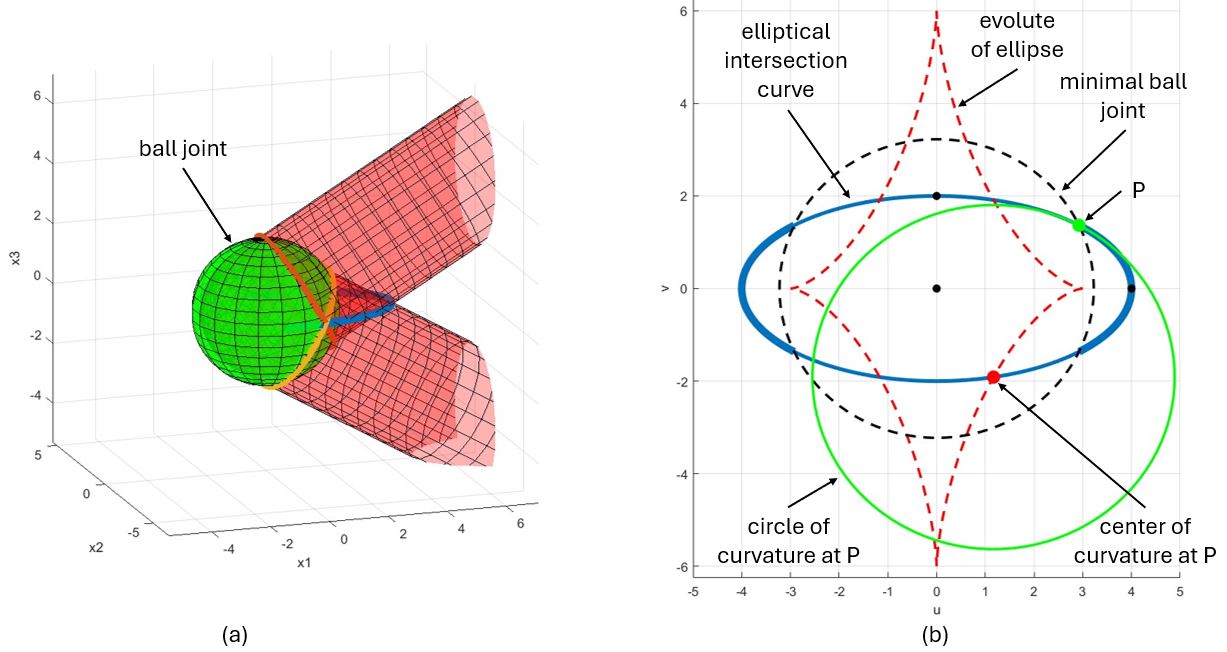}
    \caption{Illustration of a ball joint exhibiting a triple intersection point. (a) The ball joint configuration, with the elliptical intersection curve between the cylinders highlighted in blue. (b) Geometric analysis of the intersection curve, showing the evolute of the ellipse in red and the minimal enclosing ball joint in black.}
    \label{fig:ball_joint_min_radius}
\end{figure}

\textbf{Ball joint radius analysis:} While the previously described method for determining the radius of a ball joint is straightforward, it imposes a significant constraint on the available free space. This limitation can be alleviated by further analyzing the triple intersection formed by the the protrusion of the cylinders intersection curve from the sphere, as depicted in Fig.~\ref{fig:ball_joint_min_radius}(a). It is evident from Fig.~\ref{fig:ball_joint_min_radius}(a) that the triple intersection does not produce a concave region, and a potential escape direction lies along the pairwise intersection curve of the cylinders.

Recall that the intersection curve between two cylinders of equal radii and intersecting axes forms a planar ellipse. Furthermore, for there to be a test direction along a planar intersection curve, it is sufficient that the curve's evolute lies entirely within the curve itself. Thus, to ensure that the transformed ball joint has an escape direction, we must verify that the evolute of the intersection arc protruding from the ball joint is enclosed by the elliptical intersection curve.

We begin by expressing the major and minor axes of the ellipse, along with the expansion factor, as follows:
\begin{gather*}
    a = \frac{r_{\text{cylinder}} + R}{\sin(\theta/2)} = h \cdot b \\
    b = r_{\text{cylinder}} + R \\
    h = \frac{1}{\sin(\theta/2)} 
\end{gather*}

\noindent The parameterized form of the elliptical curve is
\begin{equation}
    \tau(\phi) = \begin{pmatrix} a \cos(\phi) \\ b \sin(\phi) \end{pmatrix}
    \label{eq:ellipse_curve}
\end{equation}

\noindent Its radius of curvature at a given $\phi$ is
\begin{equation}
    \rho(\phi) = \frac{\left( a^2 \sin^2(\phi) + b^2 \cos^2(\phi) \right)^{3/2}}{ab}
    \label{eq:ellipse_curvature}
\end{equation}

\noindent while its inward normal vector at the same point is
\begin{equation}
    N(\phi) = \frac{1}{\sqrt{a^2 \sin^2(\phi) + b^2 \cos^2(\phi)}} 
    \begin{pmatrix} -b \cos(\phi) \\ -a \sin(\phi) \end{pmatrix}
    \label{eq:ellipse_normal}
\end{equation}

\noindent Using Eqs.~(\ref{eq:ellipse_curve})–(\ref{eq:ellipse_normal}), we can derive the center of curvature as
\begin{equation}
    C(\phi) = \tau(\phi) + \rho(\phi) \cdot N(\phi) = 
    \begin{pmatrix} 
        \frac{(a^2 - b^2) \cos^3(\phi)}{a} \\
        \frac{(b^2 - a^2) \sin^3(\phi)}{b}
    \end{pmatrix} = 
    \begin{pmatrix} C_x \\ C_y \end{pmatrix}
\end{equation}

\begin{figure}[!t]
    \centering
    \includegraphics[width=0.85\linewidth]{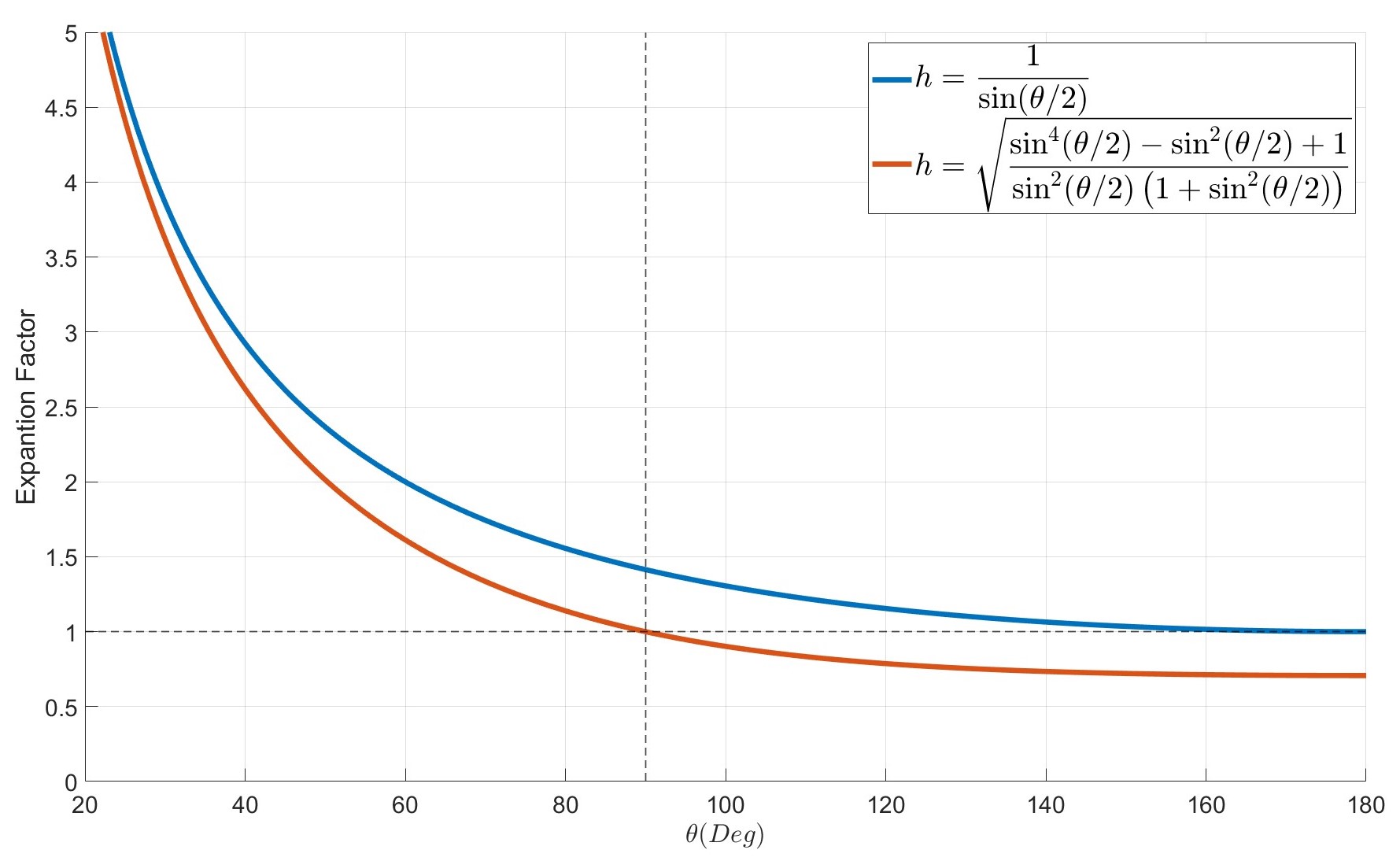}
    \caption{Comparison of the two methods for determining the expanded radius of a ball joint. The full enclosure of the intersection curve is shown in blue, while the minimal sphere method is depicted in orange. Lower expansion of the ball joint means less free space is occupied by the expanded ball joint. }
    \label{fig:ball_joint_comparison}
\end{figure}

\noindent Now using the implicit representation of the ellipse we can find the point where the evolute intersects the ellipse
\begin{equation}
    \frac{C_x^2}{a^2} + \frac{C_y^2}{b^2} = 
    \frac{(a^2 - b^2)^2 \cos^6(\phi)}{a^4} + 
    \frac{(a^2 - b^2)^2 \sin^6(\phi)}{b^4} = 1
    \label{eq:ellipse_implicit}
\end{equation}

\noindent Substituting $\cos^2(\phi) = 1 - \sin^2(\phi) = 1 - x$ and $a = h \cdot b$ into Eq.~(\ref{eq:ellipse_implicit}), we obtain the following cubic equation in $x$:
\begin{equation}
    (h^4 - 1) x^3 + 3 x^2 - 3 x + \left( 1 - \frac{h^4}{(h^2 - 1)^2} \right) = 0
\end{equation}

\noindent Solving this equation yields a single real positive root at
\begin{equation*}
    x = \frac{2h^2 - 1}{h^4 - 1} = \frac{\sin^2(\theta/2)\left(2 - \sin^2(\theta/2)\right)}{1 - \sin^4(\theta/2)}
\end{equation*}
\noindent This value of $x$ corresponds to the squared sine of the angle $\phi_k$ where the evolute first intersects the ellipse. Therefore, the corresponding angle is
\begin{equation*}
    \phi_k = \arcsin\left(\sqrt{x}\right) = \arcsin\left(\sqrt{\frac{\sin^2(\theta/2)\left(2 - \sin^2(\theta/2)\right)}{1 - \sin^4(\theta/2)}}\right)
\end{equation*}
\noindent Using this result, we can now determine the minimal radius of the ball joint such that the evolute remains entirely within the elliptical intersection curve. This radius corresponds to the norm of the position vector $\tau(\phi_k)$ and can be described as
\begin{equation}
\begin{aligned}
    r'_{\text{ball-joint}} &= \|\tau(\phi_k)\| = \sqrt{a^2 \cos^2(\phi_k) + b^2 \sin^2(\phi_k)} \\
    &= (r_{\text{cylinder}} + R) \sqrt{\frac{\sin^4(\theta/2) - \sin^2(\theta/2) + 1}{\sin^2(\theta/2) \left(1 + \sin^2(\theta/2)\right)}} \\
    &= r_{\text{ball-joint}} + R \cdot \sqrt{\frac{\sin^4(\theta/2) - \sin^2(\theta/2) + 1}{\sin^2(\theta/2) \left(1 + \sin^2(\theta/2)\right)}}=r_{\text{ball-joint}} + R \cdot h'
\end{aligned}
\label{eq:minimal_joint}
\end{equation}

Fig.~\ref{fig:ball_joint_comparison} plots a comparison of the two methods used to determine the radius of a ball joint. The blue curve represents the complete enclosure of the intersection (Eq.~(\ref{eq:fully_enclosed})), while the orange curve corresponds to the minimal ball joint method (Eq.~(\ref{eq:minimal_joint})). Notably, for $\theta \geq 90^{\circ}$ between adjacent cylinders at a ball joint, the expansion factor in the minimal radius method falls below one, indicating that the required ball joint radius is smaller than that of the cylinders. As a result, the inclusion of a ball joint joint is unnecessary in such configurations.

\textbf{Probability of successful navigation:} Both workspace transformations involve expanding the ball joint \textit{beyond} the radius of the spherical robot. Hence, there exists non-zero probability that the point-robot's initial position falls within this expanded region. Note that the ball joint expansion occupies a small portion of the spherical robot free space. In such cases, the point-mass robot would fail to reach the target since navigation cannot be initiated from within an internal obstacle. Assuming the point-robot initial position is uniformly distributed along the boundary of the workspace, the probability of failure can be expressed as
\begin{equation*}
    \text{P}_\text{s}(\text{failure}) \leq 1 - \frac{S - 4\pi\sum_{i}(r_{bj,i}')^2}{S}
\end{equation*}
\noindent where $S$ denotes the surface area of the workspace boundary after the expansion of all internal obstacles by the spherical robot radius $R$, and $r_{bj,i}'$ represents the radius of the $i$-th expanded ball joint after expansion by the radius spherical robot.

Alternatively, if the initial position is uniformly distributed throughout the free space, the failure probability is given by
\begin{equation*}
    \text{P}_\text{v}(\text{failure}) \leq 1 - \frac{V - \frac{4}{3}\pi\sum_{i}{\left((r_{bj,i} + h_iR)^3 - (r_{bj,i} + R)^3 \right)}}{V}
\end{equation*}
\noindent where $V$ is the total volume of the free space under standard expansion, $r_{bj,i}$ is the original radius (pre-expansion) of the $i$-th ball joint and $h_i$ is the expansion factor of the ball joint. It is important to note that for larger surface area $S$ and higher total volume $V$, the probability of failure is reduced.

While the inclusion of the robot's start position within the forbidden region poses a limitation for standard navigation, this issue can be mitigated. Specifically, an intermediate target can be introduced to guide the robot away from the ball joint region during the initial stage. Once the robot has cleared the critical area, the navigation can proceed using the appropriate expansion method.

%%%%%%%%%%%%%%%%%%%%%%%%%%%%%%%%%%%%%%%%%%
\subsection{Spherical Robot Simulation}
\label{sec:sphere_point_sim}
%%%%%%%%%%%%%%%%%%%%%%%%%%%%%%%%%%%%%%%%%%
\begin{figure}
    \centering
    \includegraphics[width=0.8\linewidth]{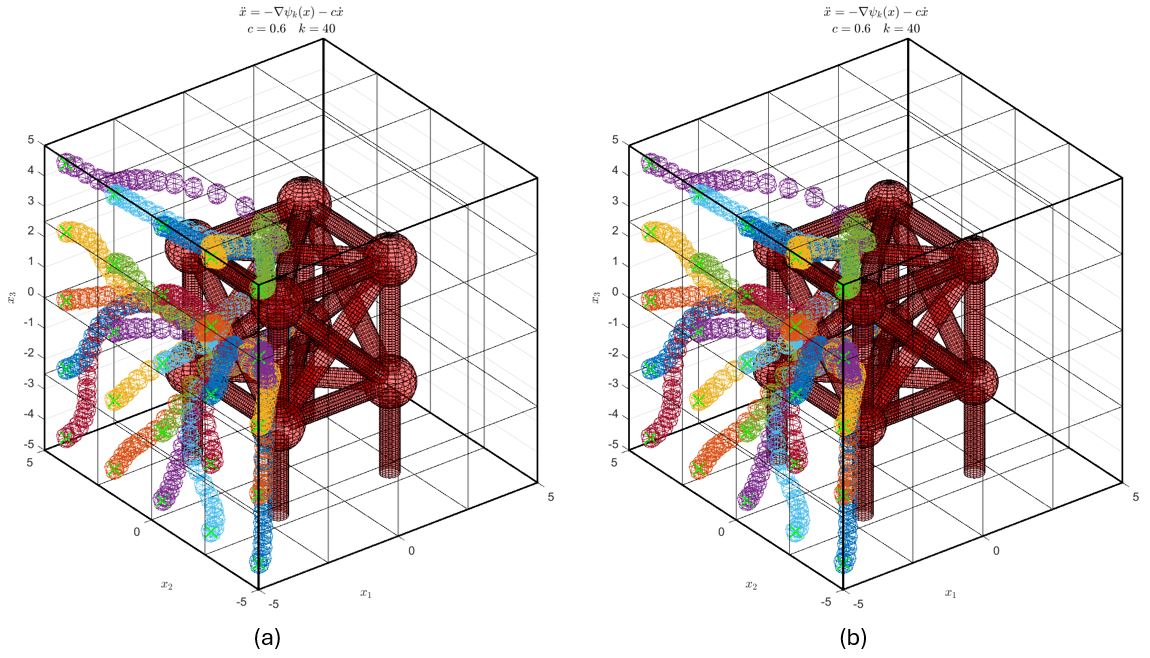}
    \caption{Simulation of a spherical robot navigating around a truss obstacle with tuning parameter $k=40$, starting from 25 initial positions with a common goal at $P_d = (0,0,0)$: (a) Trajectories generated using the full enclosure ball joint expansion method. (b) Trajectories generated using the minimal ball joint expansion method. In both cases, the spherical robot radius is set to $R = r_{\text{cylinder}}$.}
    \label{fig:ball_joint_sim}
\end{figure}

\begin{figure}[!t]
    \centering
    \includegraphics[width=0.9\linewidth]{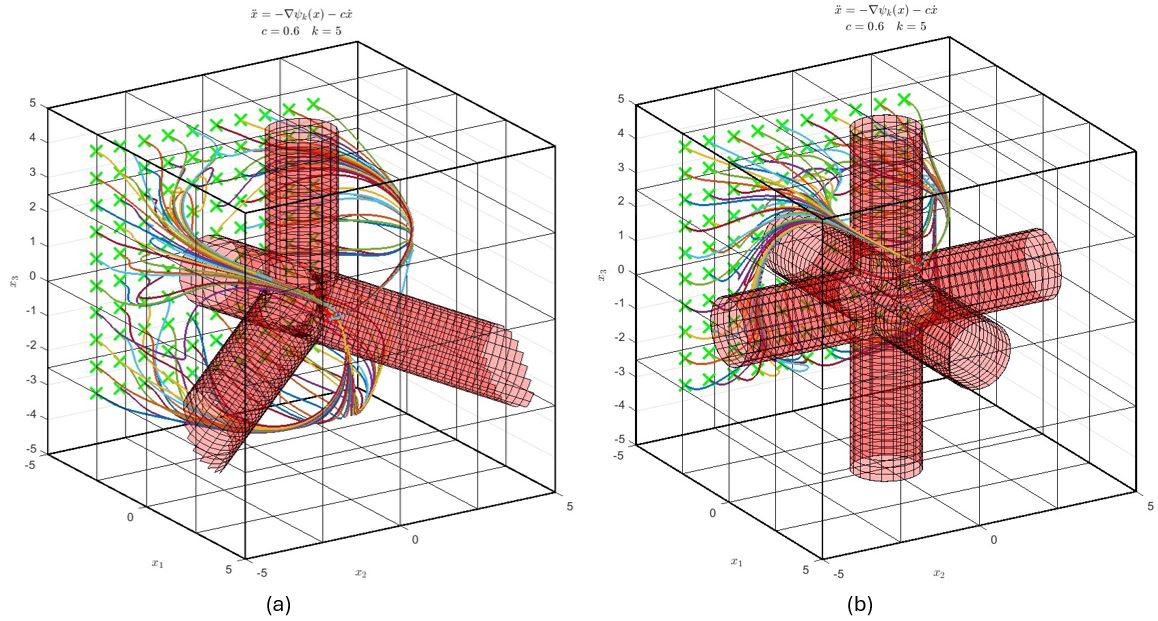}
    \caption{Spherical robot trajectories for simulation of composite obstacles with no ball joint in a cubic room for 100 initial positions and common goal. (a) Tetrahedral obstacle created using four half-cylinder obstacles, with a common goal at $P_d = (1.5,0,0)$  (b) Three mutually perpendicular full-cylinders with a common goal at $P_d = (1.5,0,1.5)$. }
    \label{fig:tetrahedron_sim}
\end{figure}

This section presents simulations of a spherical robot navigating a workspace containing the composite truss obstacle described in Section~\ref{sec:truss_sim}, using both ball joint expansion methods. Fig.~\ref{fig:ball_joint_sim}(a) shows the trajectories generated using ball joints that fully enclose the intersection curves of the cylinders. Fig.~\ref{fig:ball_joint_sim}(b) shows the trajectories for the minimal ball joint expansion method. In both simulations, the tuning parameter is set to $k=40$ and the robot radius is defined as $R = r_\text{cylinder}$. Although the minimal ball joint expansion yields a slightly larger free space, the resulting trajectories are nearly identical between the two expansion methods, indicating that both methods are viable for the motion planning problem.

\textbf{Large inter-cylinder angles:} An important observation from the minimal ball joint expansion method is that finite cylindrical obstacles with axes that start at a common point and form a pairwise angle that is greater than $90^{\circ}$ do \textit{not} require an additional spherical obstacle as a ball joint, as illustrated in Fig.~\ref{fig:tetrahedron_sim}. Fig.~\ref{fig:tetrahedron_sim}(a) shows a simulation involving a tetrahedral obstacle formation composed of four intersecting half-cylinders, each forming an angle of approximately $109^{\circ}$ with the others. Figure~\ref{fig:tetrahedron_sim}(b) presents a simulation with a composite obstacle formed by three mutually perpendicular full-cylinders. Both simulations were performed for a point robot with $k=5$, evaluating trajectories from 100 initial positions toward a common target. Since no ball joint is included in these obstacle configurations, the simulations are equivalent to that of a spherical robot, as the radii of all internal obstacles are uniformly expanded by $R$. In contrast, if ball joints were required, their required expansion would exceed $R$. Additionally, it is evident that for any reasonably small spherical robot of radius $R$, the configuration of $R$-expanded cylindrical obstacles preserves the pairwise intersections requirement, thereby maintaining the validity of the original 3-D model space. \hfill $\circ$

%%%%%%%%%%%%%%%%%%%%%%%%
%%%%%%%%%%%%%%%%%%%%%%%%
\section{Conclusions}
\label{sec:conclutions}
%%%%%%%%%%%%%%%%%%%%%%%%
%%%%%%%%%%%%%%%%%%%%%%%%

This technical report has presented a comprehensive framework for the construction and analysis of polynomial navigation functions in 3-D workspaces. The primary contributions of this work include the development of a unified polynomial framework for encoding 3-D obstacles, including spherical and various forms of capped cylindrical geometries, such as full, half, and finite cylinders. Through comprehensive gradient and Hessian analysis, we established the conditions under which these functions satisfy the fundamental properties of \textit{polarity} (a unique global minimum at the target) and \textit{admissibility} (maximal value at the free space boundary). Furthermore, we extended the framework to handle pairwise intersecting obstacles, utilizing smooth composition techniques and $p$-Rvachev functions to merge intersecting geometries while preserving the safety and smoothness of the navigation function.

The analytical results confirm that, for a sufficiently large tuning parameter $k$, the navigation function effectively eliminates local minima within the interior of the free space. By partitioning the free space into distinct regions, we proved that critical points near internal obstacles and workspace boundaries are non-degenerate saddle points rather than local minima. These theoretical results were further validated through numerical simulations in complex 3-D environments.

While this report establishes the theoretical foundations for 3-D polynomial navigation functions, several areas remain for future exploration. These include developing efficient real-time implementation strategies for robotic systems with higher degrees of freedom, investigating environment decomposition strategies for large-scale planning in complex non-convex environments, and generalizing the polynomial encoding to a broader class of obstacles beyond the current spherical and cylindrical catalog.

%%%%%%%%%%%%%%%%%%%%%%%%%%%%%
%%%%%%%%%%%%%%%%%%%%%%%%%%%%%
\bibliographystyle{ieeetr} 
\bibliography{references}
\end{document}